\documentclass{article}

\usepackage{microtype}
\usepackage{graphicx}
\usepackage{subcaption}
\usepackage{booktabs} 

\usepackage[pagebackref=false,breaklinks=true,colorlinks,bookmarks=false]{hyperref}
\hypersetup{colorlinks=true,
    linkcolor={red},
    citecolor={hanpurple},
    urlcolor={magenta}
}
\usepackage[table]{xcolor}

\usepackage{algorithm}
\usepackage{algorithmic}  

\renewcommand{\algorithmiccomment}[1]{\hfill{\footnotesize$\triangleright$~#1}}
\newcommand{\contact}{\textsuperscript{\normalfont$\star$}}
\newcommand{\corr}{\textsuperscript{\normalfont\dag}}
\usepackage{fontawesome}

\usepackage[preprint]{icml2026}
\makeatletter
\renewcommand{\printAffiliationsAndNotice}[1]{%
  \global\icml@noticeprintedtrue
  {\let\thefootnote\relax\footnotetext{%
    \noindent\hspace*{-1.9em}%
    {\scriptsize
    \begin{tabular}{@{}l@{}}
      \contact This work was done while Zichao Zeng (\href{mailto:zichao.zeng.21@ucl.ac.uk}{zichao.zeng.21@ucl.ac.uk}) was a visiting \\ student at Karlsruhe Institute of Technology. \\
      \corr Correspondence:
      \href{mailto:junwei.zheng@kit.edu}{junwei.zheng@kit.edu},
      \href{mailto:jiamingzhang@hnu.edu.cn}{jiamingzhang@hnu.edu.cn}.
    \end{tabular}%
    }%
  }}%
}
\makeatother

\usepackage{amsmath}
\usepackage{amssymb}
\usepackage{mathtools}
\usepackage{amsthm}

\usepackage{multirow}
\usepackage{arydshln}

\usepackage[capitalize,noabbrev]{cleveref}

\theoremstyle{plain}

\theoremstyle{definition}

\theoremstyle{remark}

\usepackage{pifont}

\usepackage[textsize=tiny]{todonotes}

\usepackage{tcolorbox}
\usepackage{enumitem}

\newcommand{\ms}[2]{#1 $\pm \text{\scriptsize\color{brown} #2}$}
\newcommand*\rot{\rotatebox{70}}
\newcommand{\cmark}{\ding{51}}%
\newcommand{\redbf}[1]{\textbf{\textcolor{red}{#1}}}

\def\ie{\emph{i.e.}}
\def\eg{\emph{e.g.}}

\icmltitlerunning{Faster or Stronger: Towards Flexible Visual Place Recognition via Weighted Aggregation and Token Pruning}

\begin{document}

\twocolumn[
  \icmltitle{Faster or Stronger: Towards Flexible Visual Place Recognition \\ via Weighted Aggregation and Token Pruning}







  
\begin{center}
{\bf
Zichao Zeng$^{1,2,}$\contact \quad
June Moh Goo$^{1}$ \quad
Junwei Zheng$^{2,}$\corr \quad
Weijia Fan$^{2,4}$
}

\vspace{0.06in}

{\bf
Jiaming Zhang$^{3,}$\corr \quad
Rainer Stiefelhagen$^{2}$ \quad
Jan Boehm$^{1}$
}

\vspace{0.2in}

{
$^{1}$University College London, London, UK \\
$^{2}$Karlsruhe Institute of Technology, Karlsruhe, Germany \\
$^{3}$Hunan University, Changsha, China \\
$^{4}$Shenzhen University, Shenzhen, China
}
\end{center}

\vskip 0.3in
    
\begin{center}
\begin{minipage}{0.88\textwidth}
\begin{abstract}
\vskip 0.3in
Visual Place Recognition (VPR) aims to match a query image 
to reference images of the same place in a large-scale database.
Recent state-of-the-art methods employ Vision Transformers (ViTs) as backbone foundation models to extract patch-level features that are robust to viewpoint, illumination, and seasonal variations, which are then aggregated into a compact global descriptor for retrieval.
Most existing aggregation methods uniformly pool patch tokens into learned clusters, despite the fact that different clusters often encode distinct spatial or semantic patterns and contribute unequally to VPR performance.
To address this limitation, we propose Weighted Aggregated Descriptor (\textbf{WeiAD}), which assigns weights to clusters during aggregation, producing more discriminative global representations.
Beyond accuracy, retrieval latency is a critical concern for large-scale deployments and resource-constrained edge devices.
Prior work mainly reduces latency by compressing global descriptors, while overlooking the cost of feature extraction, an issue exacerbated by ViT-based backbones. 
We therefore introduce \textbf{WeiToP}, a VPR-oriented token pruning framework that reduces feature extraction cost via self-distillation, where aggregation-induced token importance supervises a lightweight pruning module attached to an early transformer layer, enabling inference-time token pruning.
After a single joint training phase, WeiToP enables plug-and-play token pruning at inference time, allowing flexible and on-demand control over the accuracy–efficiency trade-off without additional training.
Moreover, WeiToP outperforms existing token pruning methods adapted from general vision tasks. 
\\ \faCode~The code is publicly available at~\href{https://zichaozeng.github.io/WeiToP}{https://zichaozeng.github.io/WeiToP}.
\end{abstract}
\end{minipage}
\end{center}
\vskip 0.2in
]



\printAffiliationsAndNotice{}  

\section{Introduction}\label{sec:intro}

\begin{figure}[t]
\centering
\includegraphics[width=0.9\linewidth]{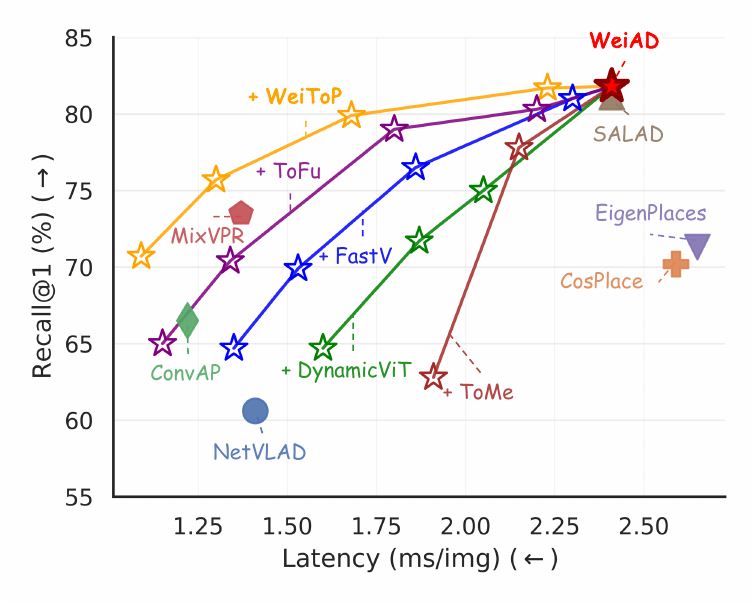}
\caption{
Star-shaped markers correspond to \textbf{WeiAD}-based models. The solid red star denotes base \textcolor{red!80!black}{\textbf{WeiAD}}. The yellow line shows our VPR-specific token pruning approach \textcolor{yellow!80!black}{\textbf{WeiToP}} integrated with WeiAD across different retention ratios. 
Other lines indicate WeiAD equipped with different generic token pruning strategies. Single markers show competing VPR methods.
}
\vskip -1em
\label{LvRALL}
\end{figure}

\begin{figure}[t]
  \centering
  \begin{subfigure}{0.35\textwidth}
    \centering
    \includegraphics[width=\linewidth]{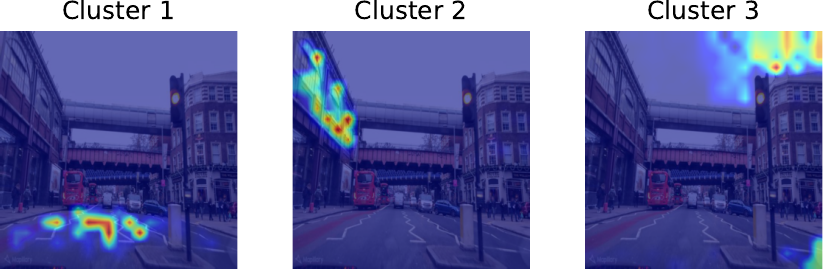}
    \caption{}\label{figa:x}
  \end{subfigure}
  \hspace{0.01\textwidth}
  \begin{subfigure}{0.10\textwidth}
    \centering
    \includegraphics[width=\linewidth]{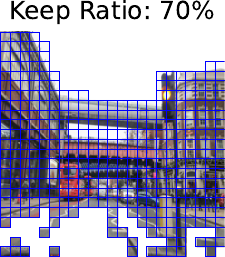}
    \caption{}\label{figb:x}
  \end{subfigure}
  \caption{
(a) Cluster-to-patch transport heatmaps showing distinct assignment patterns.
(b) Token pruning illustration, where squares denote patch tokens and blank ones are pruned redundant tokens.
  }\label{fig:x}
  \vskip -1em
\end{figure}

Visual Place Recognition (VPR) aims to retrieve previously visited locations from visual observations and serves as a core component in robotics, AR/VR, and autonomous driving systems~\cite{GSV-Cities}.
VPR pipelines typically follow a two-stage paradigm: extracting a compact global descriptor using a powerful image encoder (\eg, ResNet or Vision Transformer)~\cite{MixVPR, CosPlace, EigenPlaces, NetVLAD}, followed by efficient nearest-neighbor retrieval in large-scale databases~\cite{Faiss-gpu}.
Recent transformer-based visual foundation models, such as CLIP~\cite{CLIP} and DINOv2~\cite{DINOv2}, have further improved robustness to severe illumination, viewpoint, and seasonal variations, leading to state-of-the-art performance on challenging VPR benchmarks~\cite{AnyLoc, SegVLAD, SALAD, MuSSel-V, CricaVPR, VLAD-Buff}. 
Over the past decade, NetVLAD~\cite{NetVLAD} has served as a dominant aggregation baseline for VPR by introducing differentiable soft assignment and aggregating residuals with respect to learned cluster centers.
Building upon this paradigm, SALAD~\cite{SALAD} further improves aggregation by incorporating optimal transport (OT) formulations and explicit dustbin mechanisms to discard uninformative tokens. SuperVLAD~\cite{SuperVLAD} simplifies the aggregation by using a small number of clusters without explicit cluster centers and introduces ghost clusters similar to dustbins. 
Although effective, these approaches adopt a flat aggregation formulation: each token is either assigned to one of a fixed set of clusters or discarded altogether. 
Such designs overlook the fact that clusters which encode different spatial or semantic patterns contribute to VPR performance with varying degrees of importance (Fig.~\ref{figa:x}).

Meanwhile, efficiency is a central requirement for practical VPR systems. Most existing efficiency-oriented approaches focus on descriptor-side compression, such as dimensionality reduction, product quantization, or aggregation-level sparsification~\cite{AnyLoc, SALAD}. While these techniques reduce memory usage and retrieval cost, they leave the local feature extraction stage unchanged. In practice, feature extraction dominates the overall computational budget, affecting not only segment-level coarse retrieval~\cite{SegVLAD, MuSSel-V} and re-ranking stages~\cite{SelaVPR, R2Former}, but also large-scale deployments~\cite{ORB-SLAM,VisualSLAM,GSV-Cities,VisualNavigation} that require low latency and high throughput. Some methods offer alternatives with lighter-weight structures at the expense of performance ceilings~\cite{SALAD}. This naturally raises the question: Why not achieve a flexible balance between efficiency and accuracy by reducing redundancy in local features without sacrificing performance ceilings?

Token pruning in Vision Transformers (ViTs) has recently emerged as an effective strategy for accelerating inference by removing spatial tokens that contribute little to downstream tasks~\cite{DynamicViT, AdaViT, FitPrune, FastV}. Prior work has demonstrated its effectiveness in general vision benchmarks. Surprisingly, despite its clear potential, token pruning remains largely unexplored in ViT-based VPR, where representations are inherently global, local redundancy is substantial, and efficient feature extraction is critical (Fig.~\ref{figb:x}).

To address these limitations, we propose \textbf{Wei}ghted \textbf{A}ggregated \textbf{D}escriptor (\textbf{WeiAD}) with a \textbf{To}ken \textbf{P}runing module (\textbf{WeiToP}), forming a unified framework that jointly improves VPR robustness and efficiency. \textbf{WeiAD} extends OT-based aggregation with a VPR-oriented tiered weighting scheme and bidirectional dustbins, organizing clusters from highly informative to marginally relevant according to their contribution to place recognition. This design enables importance-weighted aggregation and yields more discriminative global descriptors. Furthermore, we introduce a self-distillation mechanism that transfers aggregation-induced token importance to a lightweight pruning module \textbf{WeiToP} attached to an early transformer layer. During inference, \textbf{WeiToP} prunes less informative tokens based on predicted importance, substantially accelerating feature extraction while preserving VPR-relevant semantics. Our main contributions are summarized as follows:

\begin{itemize}[itemsep=0pt, topsep=0pt, nosep]
\item \textbf{Weighted OT-based aggregation for VPR.} We propose \textbf{WeiAD}, a weighted optimal transport aggregation framework that assigns importance-aware weights to clusters, achieving state-of-the-art performance on challenging VPR benchmarks.
\item \textbf{VPR-oriented token pruning via self-distillation.} We introduce \textbf{WeiToP}, the first token pruning framework tailored for VPR, which distills aggregation-induced token importance into a lightweight module for inference-time pruning, reducing computation while preserving VPR-relevant semantics.
\item \textbf{Flexible accuracy–efficiency trade-off with a single training.} \textbf{WeiAD} and \textbf{WeiToP} are jointly trained once and can be used in a plug-and-play manner at inference time, enabling flexible control over the accuracy–efficiency trade-off and outperforming token pruning methods adapted from general vision tasks.
\end{itemize}


\begin{figure*}[t]
\centering
\includegraphics[width=\linewidth]{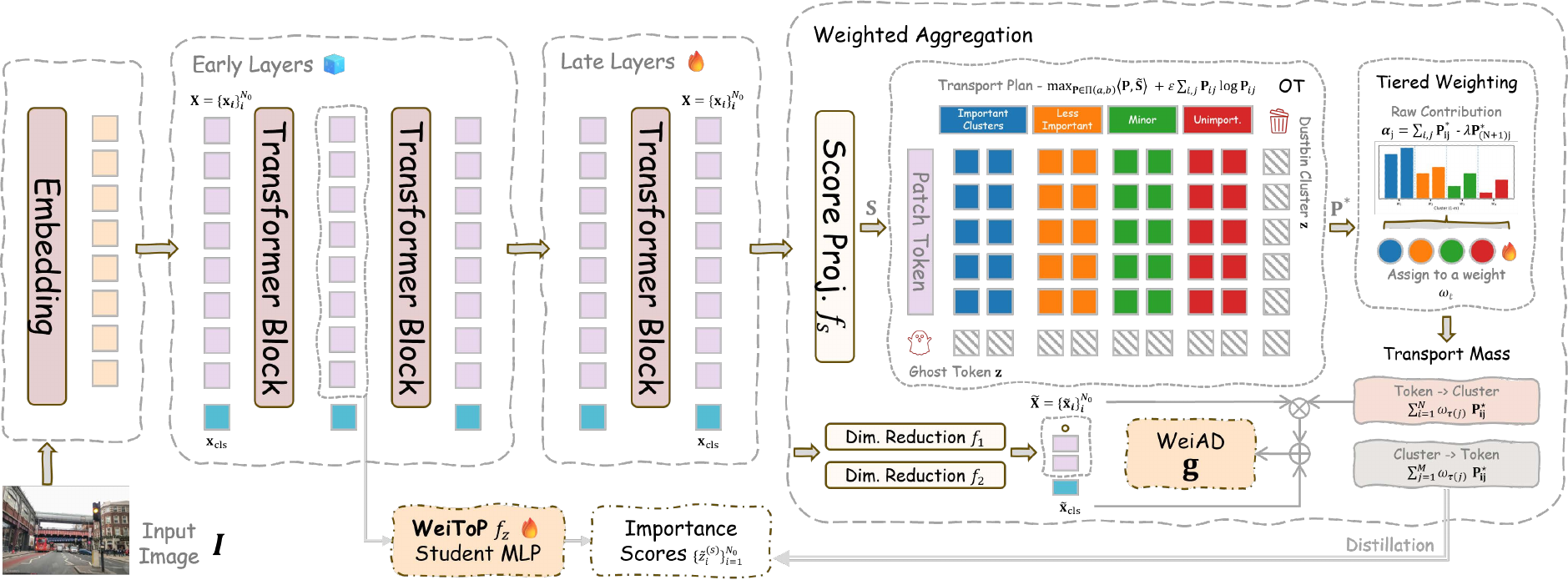}
\caption{The unified framework of \textbf{WeiAD} and \textbf{WeiToP}. At training stage, we fine-tune the late layers of DINOv2 ViT-B on GSV-Cities, alongside the score projection, dimension reduction module, WeiToP module, and weight parameters (fire icon).}\label{fig:framework}
\vskip -1em
\end{figure*}

\section{Literature Review}
\paragraph{Visual Place Recognition.}
NetVLAD~\cite{NetVLAD} marked a milestone by integrating convolutional neural networks with a differentiable VLAD aggregation layer, enabling end-to-end training for VPR. Generalized Mean Pooling (GeM)~\cite{GEM} further simplified global aggregation while maintaining competitive performance. As VPR-specific datasets became available, subsequent works reformulated VPR as a classification problem to enable scalable training. CosPlace~\cite{CosPlace} and EigenPlaces~\cite{EigenPlaces} introduced VPR-oriented training paradigms that learn highly discriminative global descriptors. MixVPR~\cite{MixVPR} further enhanced aggregation by combining deep features with MLPs. While these CNN-based approaches achieve strong performance, they typically rely on fully trained backbones and are limited in robustness under severe appearance changes. More recently, robust visual foundation models such as DINOv2~\cite{DINOv2} have emerged as powerful local feature extractors for VPR. In particular, ViT variants are well suited for VPR due to their ability to model long-range dependencies and capture global context. AnyLoc~\cite{AnyLoc} demonstrates that DINOv2 features can be directly applied to VPR without task-specific training. SALAD~\cite{SALAD} further advances this direction by formulating token-to-cluster assignment as an OT problem and introducing a dustbin cluster to absorb uninformative tokens, establishing a strong baseline with minimal fine-tuning. Similarly, SuperVLAD~\cite{SuperVLAD} fine-tunes ViT-based DINOv2 and proposes a compact single-cluster formulation with ghost clusters. Beyond global descriptor retrieval, several works explore segment-level or two-stage VPR pipelines. SegVLAD~\cite{SegVLAD} and MuSSel-V\cite{MuSSel-V} adopt DINOv2 as the backbone and perform coarse retrieval over image segments, trading efficiency for improved robustness. SelaVPR~\cite{SelaVPR} proposes a hybrid global–local adaptive framework that integrates DINOv2 into a two-stage VPR system. While these approaches are complementary to our work, the improved efficiency and representation quality of our feature extraction can naturally benefit such pipelines.

\paragraph{Token Pruning.}
Efficiency is a critical requirement for large-scale VPR systems, particularly in downstream applications operating under strict latency and power constraints~\cite{EigenPlaces}. Most existing efficiency-oriented methods focus on descriptor-level optimizations, such as dimensionality reduction~\cite{SALAD,AnyLoc} or aggregation-side sparsification~\cite{SuperVLAD}, while leaving the feature extraction process unchanged. However, with the increasing adoption of ViT-based foundation models, feature extraction has become the dominant computational bottleneck due to the large number of redundant patch tokens. Token pruning has recently emerged as an effective strategy for accelerating transformer-based models by dynamically removing tokens that contribute little to downstream tasks. This line of research has been explored in both NLP and computer vision, with pruning criteria based on attention scores~\cite{tp_EVIT_attentionscore}, token similarity~\cite{ToFu,ToMe}, or learned importance predictors~\cite{DynamicViT,DynamicPlainViTs,LTP}. More recently, numerous pruning methods requiring no training have also been explored~\cite{FitPrune,FastV,ATS,ZeroTPrune}. Despite its clear relevance, token pruning has not yet been investigated in the context of VPR. Unlike recognition tasks that focus on localized semantics, VPR requires globally consistent representations while tolerating substantial local redundancy. This makes VPR particularly well-suited for importance-guided token reduction. Our work bridges this gap by introducing a VPR-oriented token pruning strategy that leverages aggregation-induced token importance, enabling efficient feature extraction without compromising place recognition performance.

\section{Method}
\subsection{Preliminary}
\paragraph{Problem Statement.}
VPR aims to retrieve the most likely reference image depicting the same place as a given query image. Given a query image $I^{(q)}$ and a reference database $\mathcal{D} = \{ I^{(r)}_k \}_{k=1}^R$, where $R$ is the number of references, a typical pipeline consists of three stages. First, an image encoder $F(\cdot)$ extracts a set of patch-level features $\mathbf{X} = \{ \mathbf{x}_i \}_{i=1}^N$, where $\mathbf{x}_i \in \mathbb{R}^d$ denotes the $i$-th patch token embedding, and a global \texttt{[CLS]} token $\mathbf{x}_\text{CLS}$. Second, all tokens are aggregated into a global descriptor $\mathbf{g} \in \mathbb{R}^D$ tailored for VPR. Finally, retrieval is performed by computing a similarity metric (\eg, cosine or Euclidean distance) between the query descriptor $\mathbf{g}^{(q)}$ and reference descriptors $\{ \mathbf{g}^{(r)}_k \}_{k=1}^M$.

\paragraph{Local Feature Extraction.}
We adopt a ViT as the image encoder due to its ability to model global context via self-attention and its robustness across diverse visual conditions.
Given an input image $I \in \mathbb{R}^{H \times W \times C}$, it is partitioned into non-overlapping patches of size $p \times p$, where $p = 14$. This results in an initial set of $N_0 = HW / p^2$ patch tokens. The patch tokens, together with a global \texttt{[CLS]} token, are processed by a ViT consisting of $L$ transformer layers. Without token pruning, the encoder outputs
\begin{equation}
    F(I) = \{\mathbf{x}_i\}_{i=1}^{N_0} \cup \{\mathbf{x}_\text{CLS}\},
\end{equation}
where $\mathbf{x}_i \in \mathbb{R}^d$. When token pruning is applied, the number of tokens decreases after layer $\ell$. Let $N_\ell$ denote the number of patch tokens at layer $\ell$. We assume a fixed retention ratio $\rho \in (0,1]$, such that $N_\ell \approx \rho N_{\ell-1}$, where pruning is applied only once in \textbf{WeiToP}. Accordingly, $N \approx \rho N_0$, where $N$ denotes the number of final output tokens. The pruning strategy is detailed in Sec.~\ref{sec:token_pruning}.

\paragraph{Feature Aggregation and Retrieval.}
The extracted patch tokens are aggregated into a compact global descriptor $\mathbf{g} = \mathcal{A}(\mathbf{X})$, where $\mathcal{A}(\cdot)$ denotes a feature aggregation operator. In \textbf{WeiAD}, we focus on OT-based aggregation, which is introduced in Sec.~\ref{sec:weighteda}. Given a query descriptor $\mathbf{g}^{(q)}$ and a reference descriptor $\mathbf{g}^{(r)}$, retrieval is performed using a distance metric
\begin{equation}
    d(\mathbf{g}^{(q)}, \mathbf{g}^{(r)}) = \| \mathbf{g}^{(q)} - \mathbf{g}^{(r)} \|_2^2.
\end{equation}
The top-$K$ retrieved reference images are obtained as
\begin{equation}
    \pi^{(q)}_K = \arg\min_{i_1,\dots,i_K}
    \; d(\mathbf{g}^{(q)}, \mathbf{g}^{(r)}_1), \dots, d(\mathbf{g}^{(q)}, \mathbf{g}^{(r)}_K).
\end{equation}

\subsection{Weighted OT-based Aggregation}\label{sec:weighteda}
\subsubsection{OT-based Aggregation}
The framework of our methods is shown in Fig.\ref{fig:framework}. As in SALAD~\cite{SALAD}, we formulate feature aggregation as an entropy-regularized OT problem that softly assigns patch tokens to a set of clusters, enforcing bidirectional constraints between tokens and clusters, which results in more balanced assignments. Sinkhorn Algorithm~\cite{SinkHorn, SinkHornOT} is applied for OT assignment. Different to SALAD, we introduce $M$ learnable clusters together with bidirectional dustbins, \ie, a dustbin cluster and a ghost token, yielding $(N+1)$ source nodes and $(M+1)$ target nodes. A transport score matrix $\mathbf{S} = f_s(\mathbf{X}) \in \mathbb{R}^{N \times M}$ is learned, where $\mathbf{S}_{ij}$ measures the affinity between patch token $i$ and cluster $j$. Following SuperGlue~\cite{SuperGlue}, we introduce a learnable dustbin score $\mathbf{z} \in \mathbb{R}$ to handle unmatched elements. Specifically, we augment the score matrix by adding both a dustbin cluster and a ghost token, yielding an extended score matrix
\begin{equation}
\tilde{\mathbf{S}} =
\begin{bmatrix}
\mathbf{S} & \mathbf{z}\,\mathbf{1}_{N} \\
\mathbf{z}\,\mathbf{1}_{M}^\top & \mathbf{z}
\end{bmatrix}
\in \mathbb{R}^{(N+1)\times(M+1)}.
\end{equation}
The additional row corresponds to a dustbin cluster that absorbs unmatched tokens, while the additional column represents a ghost token that allows clusters with no corresponding tokens to remain unassigned. Both extensions share the same learnable score $\mathbf{z}$, ensuring a symmetric and parameter-efficient treatment of unmatched mass.

The OT plan $\mathbf{P}\in\mathbb{R}_+^{(N+1)\times(M+1)}$ is obtained by solving
\begin{equation}
\label{eq:ot}
\max_{P\in\Pi(a,b)} \;\langle \mathbf{P}, \tilde{\mathbf{S}}\rangle + \varepsilon \sum_{i,j} \mathbf{P}_{ij}\log \mathbf{P}_{ij},
\end{equation}
where $\varepsilon>0$ is the entropy regularization coefficient. This formulation is equivalent to the classical entropy-regularized OT problem with cost matrix $\mathbf{C}$ by setting $\tilde{\mathbf{S}} = -\mathbf{C}$, but is more convenient for similarity-based matching, as it directly maximizes affinities rather than minimizing costs. Finally, a factorized solution is modeled: 
\begin{equation}
\mathbf{P}^\star = \operatorname{diag}(\mathbf{u})\,K\,\operatorname{diag}(\mathbf{v}), \qquad
K_{ij}=\exp\!\left(\frac{\tilde{\mathbf{S}}_{ij}}{\varepsilon}\right),
\end{equation}
where $\mathbf{u}\in\mathbb{R}^{N+1}_+$ and $\mathbf{v}\in\mathbb{R}^{M+1}_+$ are scaling vectors obtained via the Sinkhorn algorithm.

\subsubsection{Weighted Aggregation}
We first compress patch tokens and the \texttt{[CLS]} token into spaces with lower dimensionalities: $ \tilde{\mathbf{x}}_i = f_1(\mathbf{x}_i)$ and $\tilde{\mathbf{x}}_{\texttt{CLS}}=f_2(\mathbf{x}_{\texttt{CLS}})$, where $f_1$ and $f_2$ are lightweight MLPs.

Unlike aggregation methods that treat all clusters equally, we estimate the relative importance of each cluster based on its effective mass transported from real patch tokens. Specifically, for cluster $j$, we define its raw contribution score as
\begin{equation}
\label{eq:cluster_importance}
\alpha_j = \sum_{i=1}^{N} \mathbf{P}^\star_{ij} \;-\; \lambda\, \mathbf{P}^\star_{(N+1)j},
\end{equation}
where $\mathbf{P}^\star_{ij}$ denotes the OT mass from patch token $i$ to cluster $j$, and $\mathbf{P}^\star_{(N+1)j}$ represents the mass assigned from the ghost token to cluster $j$. The hyperparameter $\lambda>0$ controls the penalty on ghost assignments. This formulation favors clusters that are strongly supported by real tokens while suppressing those dominated by ghost mass.

Clusters are ranked in descending order according to their contribution scores $\{\alpha_j\}$ and partitioned into $T^{(w)}$ rank-based tiers with fixed sizes $\{c_0, c_1, \dots, c_{T^{(w)}-1}\}$, where $\sum_{t=0}^{T^{(w)}-1} c_t = M$. Each tier $t$ is associated with a shared learnable weight $w_t$. To encourage a monotonic importance hierarchy, we parameterize the tier weights as a non-increasing sequence:
\begin{equation}
w_t = \max\!\left(
\exp(\theta_0) - \sum_{k=1}^{t} \operatorname{softplus}(\theta_{\Delta,k}),
\;\delta
\right),
\end{equation}
where $t \in \{0, \dots, T^{(w)}-1\}$. Here, $\theta_0$ and $\{\theta_{\Delta,k}\}$ are learnable parameters, and $\delta$ is a small constant for numerical stability. Each cluster $j$ inherits the weight $w_{\tau(j)}$ of its assigned tier $\tau(j)$, which is determined by its rank in the sorted list.

The final aggregated representation for cluster $j$ is computed and the global descriptor is formed by concatenating all weighted cluster descriptors together with the projected \texttt{[CLS]} token:
\begin{equation}\label{eq:globaldescriptor}
\mathbf{g} = \{\mathbf{v}_j\}^M \cup \{\tilde{\mathbf{x}}_{\texttt{CLS}}\}, \quad \mathbf{v}_j = \sum_{i=1}^{N} w_{\tau(j)}\,\mathbf{P}^\star_{ij}\,\tilde{\mathbf{x}}_i,
\end{equation}

\begin{figure}[t]
\centering
\includegraphics[width=\linewidth]{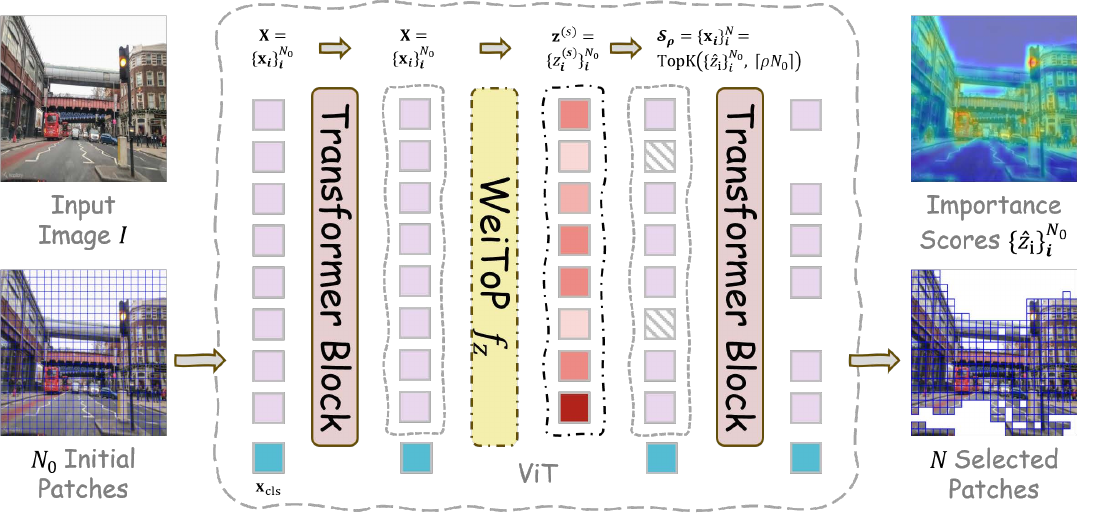}
\caption{We activate \textbf{WeiToP} after an early layer during inference. Input tokens undergo \textbf{WeiToP} processing to obtain importance logits, which are then combined with token norms to compute importance scores. The top-$\alpha$ selected tokens are retained and fed into the subsequent blocks.}
\vskip -1em
\end{figure}

\subsection{VPR-Oriented Token Pruning}\label{sec:token_pruning}

\subsubsection{Token Importance Self-Distillation}

Although weighted OT-based aggregation improves representation quality, the computational cost of extracting a large number of patch tokens remains a major bottleneck for efficient VPR. To accelerate feature extraction, we aim to prune less informative patch tokens at an early layer of the ViT backbone during inference, while preserving the semantic cues critical for VPR. According to Eq.~\ref{eq:globaldescriptor}, the contribution $\mathcal{I}_i$ of a patch token $\mathbf{x}_i$ to VPR is explicitly captured by its aggregated transport mass:
\begin{equation}
\label{eq:token_importance}
\mathcal{I}_i = \sum_{j=1}^{M} w_{\tau(j)}\,\mathbf{P}^\star_{ij},
\end{equation}
which reflects how strongly token $i$ participates in aggregation. We treat $\mathcal{I}_i$ as a soft supervision signal that encodes token-level importance for VPR.

To enable early pruning, we distill this aggregation-induced importance from the final aggregation stage to an early layer of the ViT. Specifically, we introduce a lightweight token importance predictor, termed \textbf{WeiToP}, implemented as a shared MLP $f_z$ operating on early-layer token embeddings. During training, \textbf{WeiToP} predicts importance logits $\mathbf{z}^{(s)}=\{z^{(s)}_i\}_{i=1}^{N_0}$ from the first transformer block, while the aggregation-derived importance $\mathcal{I}_i$ serves as the teacher signal. Following~\citet{distillation}, we define softened probability distributions using a temperature $T$:
\begin{equation}
p_i^{(t)} = \frac{\exp\!\left(\mathcal{I}_i / T\right)} {\sum_{k=1}^{N_0} \exp\!\left(\mathcal{I}_k / T\right)}, \quad
p_i^{(s)} = \frac{\exp\!\left(z^{(s)}_i / T\right)} {\sum_{k=1}^{N_0} \exp\!\left(z^{(s)}_k / T\right)}.
\end{equation}

The distillation objective encourages the student predictor to match the teacher distribution:
\begin{equation}
\label{eq:distill_loss}
\mathcal{L}_{\mathrm{distill}} = T^2 \cdot \frac{1}{N_0} \sum_{i=1}^{N_0} p_i^{(t)} \log \frac{p_i^{(t)}}{p_i^{(s)}}.
\end{equation}
Notably, WeiToP is trained jointly with the full VPR pipeline but remains inactive during training, ensuring that pruning does not interfere with representation learning.

\subsubsection{Flexible Pruning during Inference}
At inference time, WeiToP is activated to enable efficient token pruning. Given early-layer patch token embeddings, WeiToP predicts token-wise importance logits $\{ z^{(s)}_i \}_{i=1}^{N_0}$, which reflect their estimated relevance to VPR. To obtain a comparable importance scale, the logits are first normalized using min--max normalization. In addition to the learned importance, we incorporate the magnitude of token embeddings as a complementary saliency prior. Specifically, the final pruning score for each token is defined as a convex combination of the normalized importance and the token embedding norm:
\begin{equation}
\hat{z}_i = \kappa \, \frac{z^{(s)}_i - \min_j z^{(s)}_j} {\max_j z^{(s)}_j - \min_j z^{(s)}_j + \epsilon} + (1-\kappa)\, \left\lVert x_i \right\rVert_2,
\end{equation}
where $\epsilon$ is a small constant for numerical stability, $x_i$ denotes the corresponding patch token embedding and $\kappa \in [0,1]$ controls the balance between learned importance and feature magnitude. Tokens are ranked according to $\{\hat{z}_i \}$, and only the top-$\rho$ fraction is retained:
\begin{equation}
\mathcal{S}_\rho = \operatorname{TopK}\!\left(\{ \hat{z}_i \}_{i=1}^{N_0},\;\lceil \rho N_0 \rceil\right),
\end{equation}
where $\rho \in (0,1]$ controls the pruning ratio. Only the selected tokens $\mathcal{S}_\rho$ are propagated through the remaining transformer layers and aggregated by WeiAD. This design enables a flexible efficiency–accuracy trade-off at inference time without retraining.


\subsection{Loss Function}
The overall training objective consists of a retrieval loss and a token importance distillation loss.
We adopt the multi-similarity loss~\cite{MultiSimLoss} to supervise global descriptor learning. Given a query descriptor $\mathbf{g}_i$, positive set $\mathcal{P}_i$, and negative set $\mathcal{N}_i$, the loss is defined as
\begin{equation}
\begin{split}
\mathcal{L}_{\mathrm{retr}}=
&\frac{1}{\alpha}\log\!\left(1 + \sum_{p \in \mathcal{P}_i} \exp\!\left(-\alpha (s_{ip} - \lambda)\right) \right)
\\
&+ \frac{1}{\beta} \log\!\left(1 + \sum_{n \in \mathcal{N}_i} \exp\!\left(\beta (s_{in} - \lambda)\right) \right),
\end{split}
\end{equation}
where $s_{ij}$ denotes the cosine similarity between descriptors, and $\alpha$, $\beta$, and $\lambda$ are hyperparameters.
\paragraph{Overall objective.}
The final training objective is
\begin{equation}
\mathcal{L}_{\mathrm{total}}=\mathcal{L}_{\mathrm{retr}}+\gamma\,\mathcal{L}_{\mathrm{distill}},
\end{equation}
where $\gamma$ balances retrieval and pruning supervision.

\begin{table*}[t]
\caption{Comparison of \textbf{WeiAD} with state-of-the-art methods on diverse VPR datasets. The best results are highlighted in \textbf{bold}, and the second-best results are \underline{underlined}. \textbf{WeiToP}, with different retention ratios $\rho$ based on \textbf{WeiAD}, is also compared. \redbf{Bold red} indicates that the performance after pruning still achieves or surpasses the best results.}\label{tab:vpr_performance}
\resizebox{\linewidth}{!}{
\begin{tabular}{llcccccccccccccc>{\columncolor{gray!10}}c>{\columncolor{gray!10}}c}
\toprule[1.5pt]
\multirow{2}{*}{\textbf{Method}} & \multirow{2}{*}{\textbf{Backbone}} & \multirow{2}{*}{$\boldsymbol{\rho}$} & \textbf{Latency} & \multicolumn{2}{c}{\textbf{MSLS-C}} & \multicolumn{2}{c}{\textbf{MSLS-val}} & \multicolumn{2}{c}{\textbf{NordLand}} & \multicolumn{2}{c}{\textbf{Pitts250k-test}} & \multicolumn{2}{c}{\textbf{SPED}} & \multicolumn{2}{c}{\textbf{AmsterTime}} & \multicolumn{2}{>{\columncolor{gray!10}}c}{\textbf{Mean}} \\ \cmidrule(lr){5-6} \cmidrule(lr){7-8} \cmidrule(lr){9-10} \cmidrule(lr){11-12}  \cmidrule(lr){13-14} \cmidrule(lr){15-16} \cmidrule(lr){17-18} 
                        &&                          &     (ms)                      & R@1          & R@5         & R@1           & R@5          & R@1           & R@5          & R@1              & R@5             & R@1         & R@5        & R@1         & R@5        & R@1         & R@5        \\ \midrule[1.5pt]
NetVLAD        & VGG-16    & /    & 1.41 & 35.1 & 47.4 & 82.6 & 89.6 & 32.6 & 47.1 & 90.5 & 96.2 & 78.7 & 88.3 & 18.4 & 32.4 & 56.3 & 66.8  \\ 
CosPlace       & VGG-16    & /    & 2.59 & 67.2 & 78.0 & 87.4 & 93.0 & 44.2 & 59.7 & 92.1 & 97.5 & 80.1 & 89.6 & 47.4 & 70.2 & 69.7 & 81.3  \\
Conv-AP        & ResNet-50 & /    & 1.22 & 54.2 & 66.6 & 83.1 & 90.3 & 42.7 & 58.9 & 92.9 & 97.7 & 79.2 & 88.6 & 34.6 & 54.4 & 64.5 & 76.1 \\
MixVPR         & ResNet-50 & /    & 1.37 & 64.0 & 75.9 & 88.0 & 92.7 & 58.4 & 74.6 & 94.6 & 98.3 & 85.2 & 92.1 & 41.2 & 59.6 & 71.9 & 82.2 \\
EigenPlaces    & ResNet-50 & /    & 2.65 & 67.4 & 77.1 & 89.3 & 93.7 & 54.4 & 68.8 & 94.1 & 98.0 & 69.9 & 82.9 & 48.8 & 68.8 & 70.7 & 81.5 \\
SuperVLAD      & DINOv2-B  & /    & 7.38 & 74.2 & 79.3 & \underline{91.5} & 95.9 & 60.6 & 76.0 & \textbf{95.2} & \underline{98.6} & \textbf{90.6} & \textbf{95.4} & 54.7 & 75.5 & 77.8 & 86.8  \\
SALAD          & DINOv2-B  & /    & 2.41 & \underline{75.4} & \underline{87.4} & 91.4 & \textbf{96.2} & \underline{71.1} & \underline{85.6} & 94.9 & 98.5 & 90.3 & \textbf{95.4} & \underline{58.0} & \underline{78.6} & \underline{80.2} & \underline{90.3} \\ \addlinespace[0.5ex]
\hdashline \addlinespace[0.5ex]

WeiAD          & DINOv2-B  & 1.00 & 2.41 & \textbf{76.2} & \textbf{87.8} & \textbf{91.7} & \textbf{96.2} & \textbf{73.1} & \textbf{86.7} & \underline{95.1} & \textbf{98.8} & \textbf{90.6} & \textbf{95.4} & \textbf{58.4} & \textbf{79.5} & \textbf{80.9} & \textbf{90.7}  \\ \addlinespace[0.5ex]
\hdashline  \addlinespace[0.5ex]
\; + WeiToP       & DINOv2-B  & 0.95 & 2.23 & \redbf{76.3} & \redbf{87.9} & \redbf{91.8} & \redbf{96.3} & 72.8 & \redbf{86.7} & 95.0 & 98.6 & 90.4 & 95.3 & \redbf{58.4} & 79.4 & 80.8 & \redbf{90.7}   \\
\; + WeiToP       & DINOv2-B  & 0.70 & 1.68 & 75.3 & 86.8 & 91.3 & 96.1 & 69.5 & 84.7 & 94.0 & 98.3 & 89.2 & 94.4 & 55.3 & 76.9 & 79.1 & 89.5 \\
\; + WeiToP       & DINOv2-B  & 0.50 & 1.30 & 71.6 & 84.2 & 90.4 & 95.9 & 62.3 & 80.0 & 91.5 & 97.4 & 85.5 & 92.5 & 48.7 & 70.6 & 75.0 & 86.8 \\
\; + WeiToP       & DINOv2-B  & 0.40 & 1.09 & 68.2 & 81.9 & 87.7 & 94.8 & 53.3 & 71.0 & 88.2 & 96.2 & 82.4 & 90.9 & 41.8 & 64.0 & 70.3 & 83.1  \\ \bottomrule[1.5pt]
\end{tabular}
}
\end{table*}

\begin{figure*}[t]
  \centering

  \begin{subfigure}{0.48\textwidth}
    \centering
    \includegraphics[width=\linewidth]{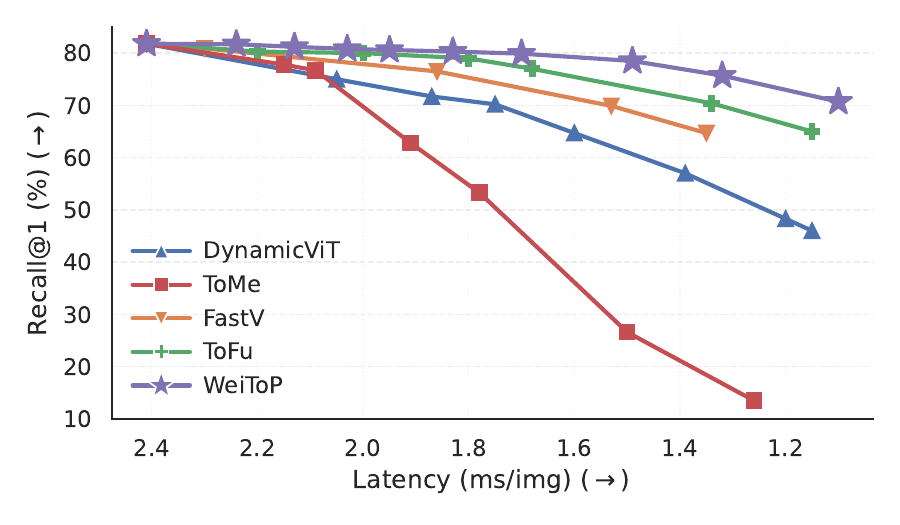}
  \end{subfigure}
  \hspace{0.01\textwidth}
  \begin{subfigure}{0.48\textwidth}
    \centering
    \includegraphics[width=\linewidth]{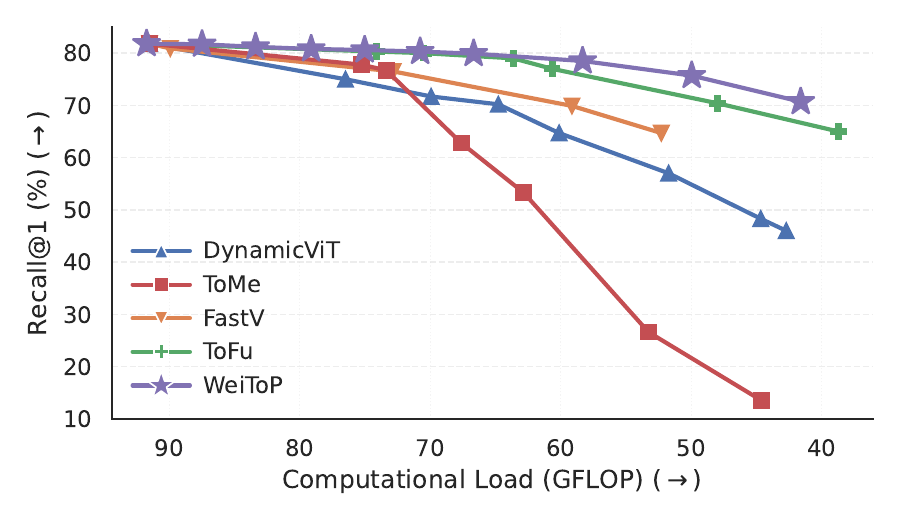}
  \end{subfigure}
  \caption{Comparison of \textbf{WeiToP} (star lines) with other token pruning methods with different retention ratios $\rho$.}\label{fig:trimmer_comparison}
  \vskip -1em
\end{figure*}

\begin{figure}
\centering
\includegraphics[width=\linewidth]{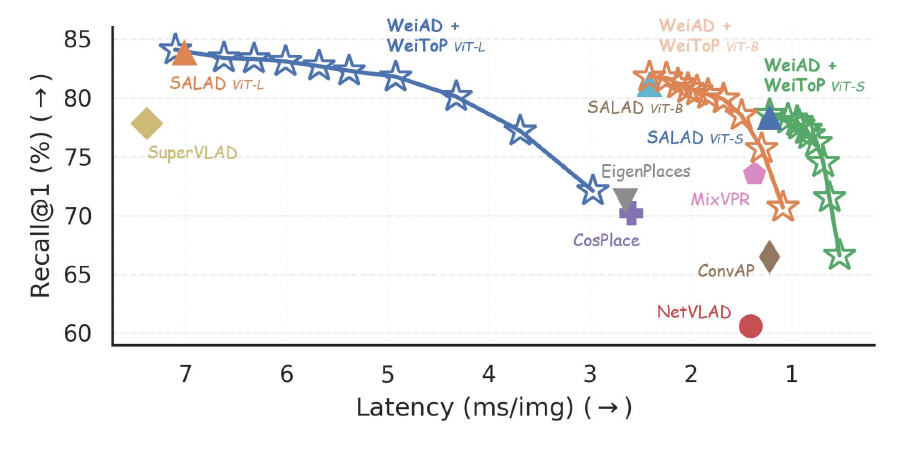}
\caption{Efficiency-accuracy trade-off performance of \textbf{WeiAD} + \textbf{WeiToP} (star lines) with different model sizes of DINOv2.}\label{fig:diff_backbone}
\label{diffsize}
\vskip -1em
\end{figure}

\section{Experiments}\label{sec:Exp}
\subsection{Implementation Details}
All experiments use PyTorch on a single RTX 3090. We adopt DINOv2 ViT-B/14 as the backbone, freezing all but the last four transformer layers. Models are trained on GSV-Cities~\cite{GSV-Cities} for 4 epochs (\~32 min). 
We use AdamW with an initial LR of $6\times10^{-5}$ and linear decay. Images are resized to $224\times224$ for training ($N_0=14\times14$) and $322\times322$ for inference ($N_0=23\times23$).
For \textbf{WeiAD}, we set $M=64$ clusters. Patch tokens are projected from 768$\rightarrow$128 dims and the \texttt{[CLS]} token from 768$\rightarrow$256, yielding a $128\times64+256$ global descriptor. A tiered weighted aggregation with $T^{(w)}=4$ tiers uses $\{24,20,16,4\}$ clusters (high$\rightarrow$low importance). The base weight is initialized as $\exp(\theta_0)=1$, weight deltas are uniform, and all tier weights are lower-bounded by $10^{-3}$. The penalty weight $\lambda$ is $0.5$.  
\textbf{WeiToP} is trained jointly but inactive. 
At inference, WeiToP predicts token importance and keeps the top-$\rho$ fraction ($\rho\in(0,1]$); pruning is enabled when $\rho<1$. The coefficient $\kappa$ is $0.5$. 
WeiADs are supervised with Multi-Similarity loss ($\alpha=1.0$, $\beta=20$, $\lambda=0$). The distillation temperature $T=0.1$ and the weighting factor weight $\gamma=0.1$.

\paragraph{Datasets and evaluation metrics.}
We evaluate \textbf{WeiAD} (+ \textbf{WeiToP}) on a diverse set of standard VPR benchmarks covering large-scale urban scenes, seasonal variation, and long-term appearance changes including \textbf{MSLS-val}idation set~\cite{MSLS}, \textbf{MSLS-C}hallenge, \textbf{Nordland} test set~\cite{Nordland}, \textbf{Pitts250k-test} set~\cite{NetVLAD},  \textbf{SPED-test} set~\cite{SPED}, and \textbf{AmsterTime} dataset~\cite{AmsterTime}. More details are in Appendix~\ref{sec:datadescri}.
We follow standard VPR protocols and report Recall@1 and Recall@5. Efficiency is measured using the R2Former~\cite{R2Former} VPR latency metric, which captures end-to-end inference time from image input to global descriptor generation, including feature extraction. We further analyze the efficiency–accuracy trade-off of WeiToP by reporting inference latency and computational load under different token retention ratios $\rho$, flexible control over accuracy and computational cost at inference.

\begin{figure*}[t]
  \centering
  \begin{subfigure}{0.282\textwidth}
    \centering
    \includegraphics[width=\linewidth]{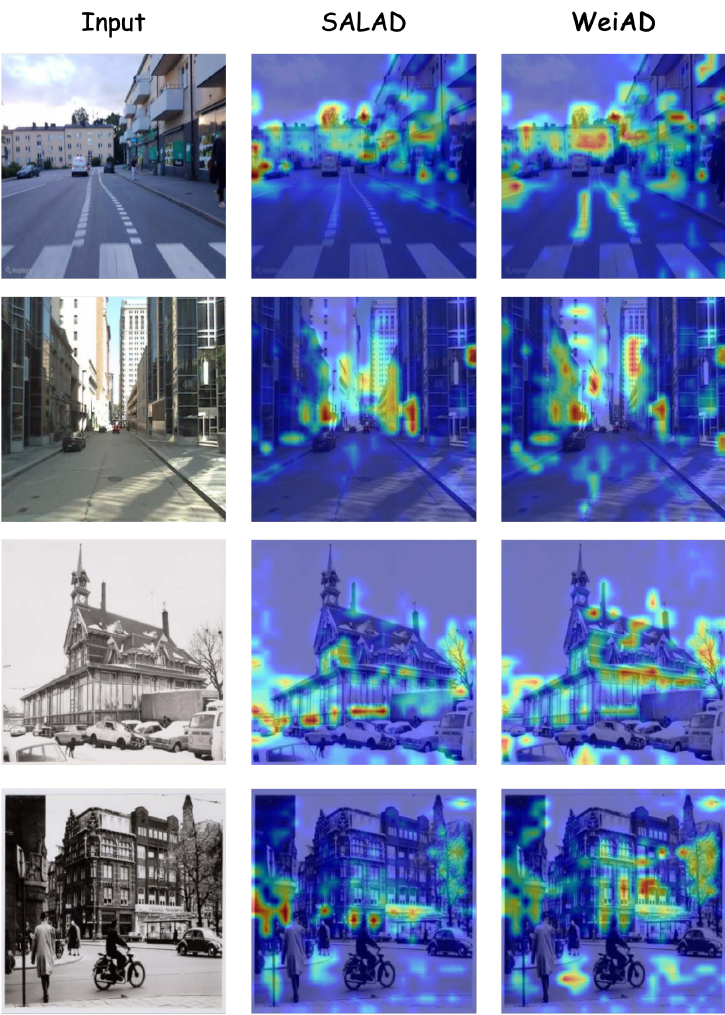}
    \caption{}\label{fig:atten}
  \end{subfigure}
  \hspace{0.02\textwidth}
  \begin{subfigure}{0.67\textwidth}
    \centering
    \includegraphics[width=\linewidth]{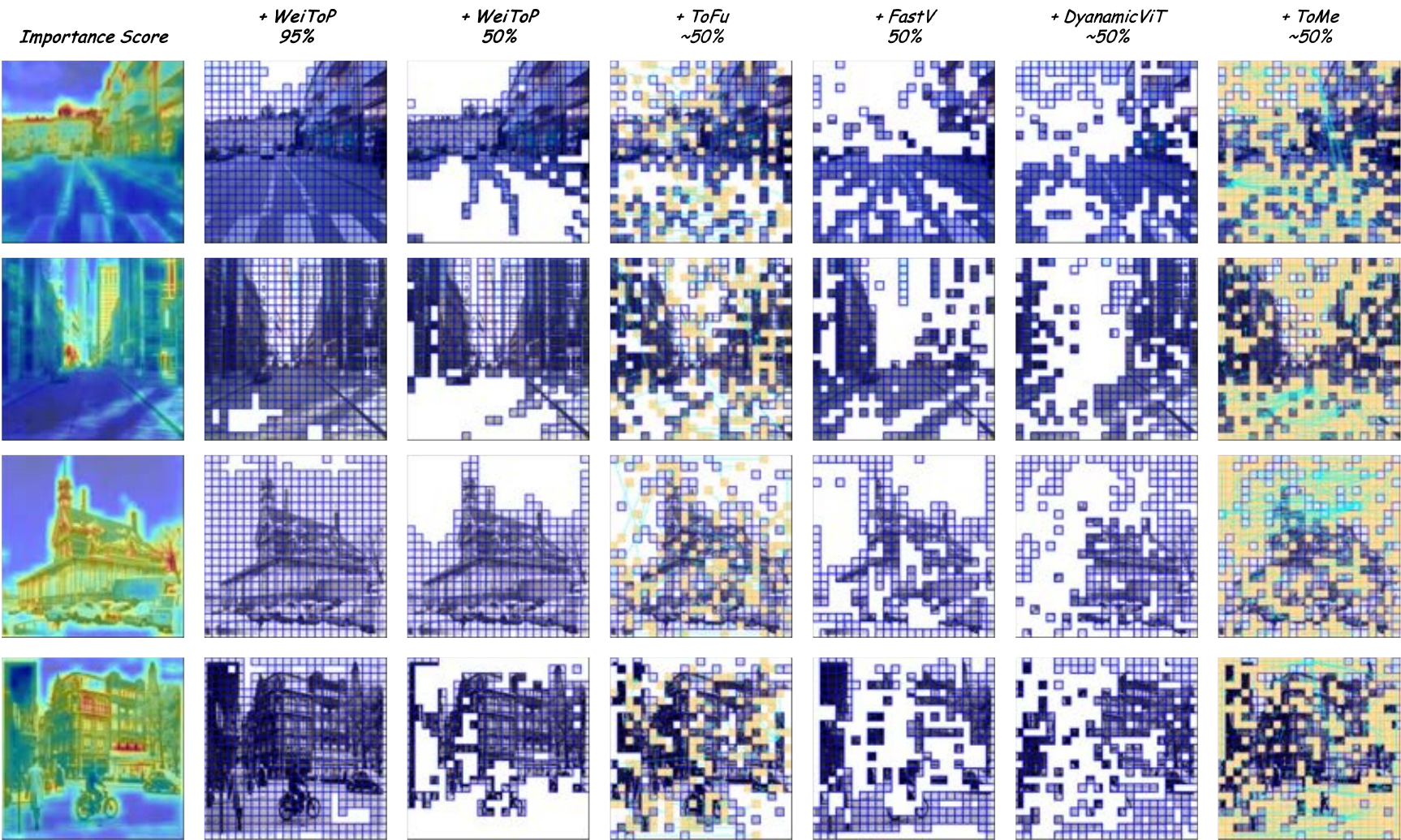}
    \caption{}\label{fig:tokenprun}
  \end{subfigure}
  \vskip -0.5em
  \caption{Visual examples under different conditions. (a) Visualization of the transport mass each token adsorbed in \textbf{WeiAD}, \ie, $\sum_{j=1}^{M} w_{\tau(j)}\,\mathbf{P}^\star_{ij}$, compared with $\sum_{j=1}^{M}\mathbf{P}^\star_{ij}$  in \textbf{SALAD}. (b) Tokens retained after applying \textbf{WeiToP} with retention ratios $\rho=0.95$ and $0.5$, compared with \textbf{ToFu}, \textbf{FastV}, \textbf{DynamicViT}, and \textbf{ToMe} at approximately $\rho=0.5$. *\textcolor{gray}{Blank} - removed tokens; \textcolor{orange}{Orange} - merged tokens.}\label{fig:visual}
  \vskip -1em
\end{figure*}

\subsection{Performance on VPR}
We benchmark \textbf{WeiAD} against a wide range of state-of-the-art single-stage VPR methods. These include representative CNN-based approaches with fully trained backbones, such as Conv-AP~\cite{GSV-Cities}, NetVLAD~\cite{NetVLAD}, CosPlace~\cite{CosPlace}, MixVPR~\cite{MixVPR}, and EigenPlaces~\cite{EigenPlaces}, as well as recent ViT-based methods built upon DINOv2 with light fine-tuning, including CricaVPR~\cite{CricaVPR}, SuperVLAD~\cite{SuperVLAD}, and SALAD~\cite{SALAD}. For a fair comparison, SALAD and our method (\textbf{WeiAD} + \textbf{WeiToP}) are each trained five times with different random seeds, and we report the averaged performance. Implementation details of all methods and the complete results are provided in Appendix~\ref{sec:implemencompare} and \ref{sec:completeresults}.

Tab.~\ref{tab:vpr_performance} summarizes the VPR performance across all evaluated benchmarks. Our method consistently outperforms prior approaches on all datasets and metrics. While performance on MSLS-val, Pitts250k-test, and SPED-test appears close to saturation for existing methods, \textbf{WeiAD} still achieves consistent improvements. More notably, on challenging benchmarks that exhibit severe seasonal and temporal variations, such as Nordland and AmsterTime, our approach yields substantial gains by leveraging more discriminative representations. On the MSLS-C benchmark, \textbf{WeiAD} achieves a clear margin over existing methods, demonstrating strong generalization under strict evaluation protocols.

\paragraph{Flexible efficiency–accuracy trade-offs.}
Beyond accuracy, we further analyze efficiency in terms of inference latency. Without token pruning, \textbf{WeiAD} exhibits the same latency as SALAD and is consistently faster than other ViT-based methods. 
CNN-based methods, particularly MixVPR and Conv-AP, remain strong baselines in terms of raw efficiency. However, when \textbf{WeiToP} is activated, our method achieves a favorable accuracy--efficiency trade-off. At a token retention ratio of $\rho=0.5$, \textbf{WeiAD}+$\,$\textbf{WeiToP} is faster than MixVPR while maintaining higher Recall@1 by $3.1\%$. When further reducing the ratio to $\rho=0.4$, our method becomes $0.29\,\mathrm{ms}$/image faster than the fastest CNN-based baseline (Conv-AP), while still improving average Recall@1 by $5.8\%$. Such improvements are particularly impactful for large-scale deployment scenarios involving thousands or millions of images, where faster global retrieval leads to multiplicative gains in end-to-end system responsiveness. Moreover, since $\rho$ is adjustable at inference time, our framework allows practical systems to \textbf{\textit{dynamically trade accuracy for efficiency}} based on deployment constraints.

\subsection{Comparison to other Token Pruning}
We further compare VPR-Oriented token pruning \textbf{WeiToP} with existing methods. To this end, we integrate several state-of-the-art generic token pruning approaches into our VPR pipeline, including DynamicViT~\cite{DynamicViT}, ToMe~\cite{ToMe}, FastV~\cite{FastV}, and ToFu~\cite{ToFu}. These methods are designed for general vision tasks and are integrated into our framework using their default configurations detailed in Appendix~\ref{sec:implemencompare2}. 

Fig.~\ref{fig:trimmer_comparison} reports Recall@1 under varying $\rho$, together with corresponding computational load and VPR latency. When $\rho$ is high, all methods exhibit comparable computational cost, and WeiToP achieves slightly higher Recall@1 than generic pruning approaches. As the $\rho$ decreases, however, generic trimmers suffer from severe performance degradation. By leveraging aggregation-induced supervision through self-distillation, WeiToP consistently achieves a superior efficiency–accuracy trade-off across different pruning regimes. Fig.~\ref{fig:tokenprun} qualitatively demonstrates the pruning rationale of \textbf{WeiToP} by discarding VPR-specific uninformative tokens compared to other methods. We additionally implemented WeiToP on DINOv2 ViT-S\&-L shown in Fig.~\ref{diffsize}. The results demonstrate WeiToP's scalability. Across models of varying sizes, flexible pruning strategies allow the model to maintain the upper bound on accuracy.


\begin{figure}[t]
\centering
\includegraphics[width=\linewidth]{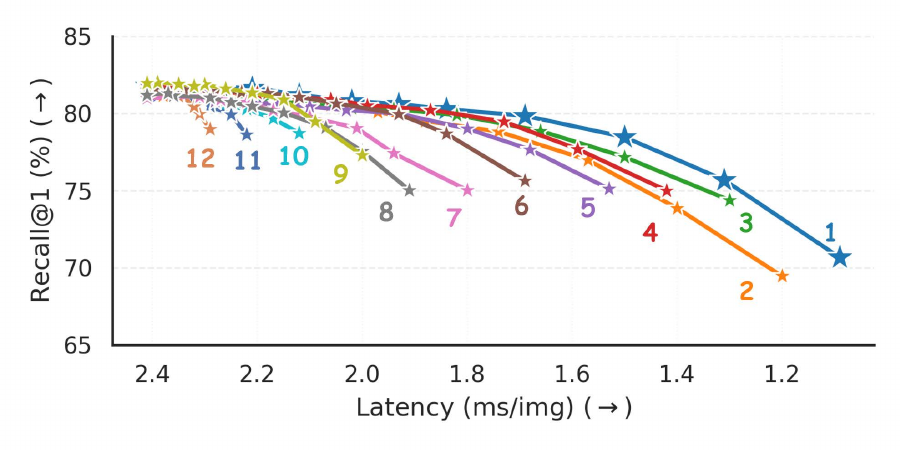}
\caption{Efficiency-accuracy trade-off of the \textbf{WeiToP} pruning module located after different layers.}\label{fig:layer}
\vskip -1em
\end{figure}

\begin{table}[t]
\caption{Effect of bidirectional dustbins and tiered weighting.}\label{tab:aba}
\centering
\resizebox{\columnwidth}{!}{
\begin{tabular}{ccc>{\columncolor{gray!10}}c>{\columncolor{gray!10}}c>{\columncolor{gray!10}}c>{\columncolor{gray!10}}c}
\toprule[1.5pt]
\textbf{Dustbin}    & \textbf{Ghost}      & \textbf{Tiered}     &   \multicolumn{4}{>{\columncolor{gray!10}}c}{\textbf{Mean} \text{\scriptsize w/o MSLS-C}}            \\ \cmidrule(lr){4-7} 
\textbf{Cluster}    & \textbf{Token}      & \textbf{Weighting}  & R@1      & R@5    & R@10   & R@20   \\ \midrule
$\times$ & $\times$ & $\times$ & \ms{81.2}{0.84} & \ms{90.8}{0.63} & \ms{\textbf{93.4}}{0.55} & \ms{\underline{95.2}}{0.37} \\ \addlinespace[0.5ex]
 \hdashline
\multicolumn{7}{l}{\textit{Bidirectional Dustbin}} \\
\cmark & $\times$ & $\times$ & \ms{81.1}{0.99} & \ms{90.9}{0.77} & \ms{93.2}{0.62} & \ms{94.9}{0.52} \\ 
$\times$ & \cmark & $\times$ & \ms{\underline{81.5}}{1.05} & \ms{91.0}{0.80} & \ms{93.1}{0.68} & \ms{94.9}{0.52} \\ 
\cmark & \cmark & $\times$ & \ms{\underline{81.5}}{0.73} & \ms{\underline{91.2}}{0.48} & \ms{93.2}{0.56} & \ms{\textbf{95.3}}{0.44} \\  \addlinespace[0.5ex]
\hdashline
\multicolumn{7}{l}{\textit{Tiered Weighting}} \\
$\times$ & $\times$ & \cmark & \ms{81.3}{1.28} & \ms{90.9}{1.07} & \ms{92.8}{0.83} & \ms{94.6}{0.61} \\  
\cmark & \cmark & \cmark & \ms{\textbf{81.8}}{0.82} & \ms{\textbf{91.3}}{0.71} & \ms{\textbf{93.4}}{0.71} & \ms{\underline{95.2}}{0.48} \\ \bottomrule[1.5pt]
\end{tabular}
}
\end{table}

\subsection{Ablation Study}

We conduct a series of ablation studies to analyze the impact of key design choices in both \textbf{WeiAD} and \textbf{WeiToP}. All ablations are performed on representative benchmarks, and detailed numerical results and the choices of hyperparameters are provided in Appendix~\ref{sec:moreablation}.

\paragraph{Effect of bidirectional dustbins and tiered weighting.}
We evaluate the roles of the dustbin cluster and ghost token in OT-based aggregation (Tab.~\ref{tab:aba}). Using only a dustbin cluster (as in SALAD) consistently improves performance over removing both. The ghost token alone also brings gains, while combining both achieves the best results, showing that bidirectional mass absorption is most effective when unmatched tokens and ambiguous clusters are jointly modeled. 
Adding importance-based cluster weighting further boosts performance, highlighting the benefit of differentiating clusters by their relevance to VPR.
Visualizations (Fig.~\ref{fig:atten}) qualitatively show that the weighted-aggregated descriptors absorb more mass from VPR-relevant regions (\eg, buildings) while suppressing irrelevant ones (\eg, roads and sky), resulting in more discriminative descriptors.

\paragraph{Location of WeiToP.}
We study where to place the pruning module in the ViT. 
Fig.~\ref{fig:layer} illustrates that placing it too late reduces the accuracy–efficiency trade-off since fewer layers benefit from pruning. We therefore place the student after the first transformer block, achieving the best balance between representation quality and pruning effectiveness.

\section{Conclusion}
In this work, we propose \textbf{WeiAD}, a \textbf{wei}ghted OT-based \textbf{a}ggregated \textbf{d}escriptor. By introducing tiered weighting into OT-based clustering, WeiAD produces \textbf{\textit{more discriminative}} global representations. To address the efficiency bottleneck of feature extraction, we further propose \textbf{WeiToP}, a VPR-oriented \textbf{to}ken \textbf{p}runing module that distills aggregation-induced token importance into early transformer layers. Unlike generic token pruning methods, WeiToP leverages task-specific supervision derived from the aggregation process, enabling \textbf{\textit{flexible and effective}} pruning at inference time. Extensive experiments on diverse VPR benchmarks demonstrate that WeiAD improves performance and that WeiToP achieves a favorable accuracy–efficiency trade-off for VPR. Notably, this work provides the first principled integration of token pruning into VPR, enabling flexible and efficient systems while maintaining the performance upper bound. Our approach offers a principled direction for VPR and can be naturally extended to cross-scenario, large-scale, multi-stage, or segment-level VPR pipelines, and combined with other efficiency-enhancing techniques such as lightweight models and dimensionality reduction.

\section*{Acknowledgment}
This work was mainly supported by the Engineering and Physical Sciences Research Council through an industrial CASE studentship with Ordnance Survey (Grant number EP/W522077/1 and EP/X524840/1). This work was supported in part by National Natural Science Foundation of China under Grant No. 62503166, in part by Helmholtz Association of German Research Centers, in part by the Ministry of Science, Research and the Arts of Baden-W\"urttemberg (MWK) through the Cooperative Graduate School Accessibility through AI-based Assistive Technology (KATE) under Grant BW6-03, and in part by the Helmholtz Association Initiative and Networking Fund on the HAICORE@KIT and HOREKA@KIT partition. 

\bibliography{main}
\bibliographystyle{icml2026}

\newpage
\appendix
\onecolumn

\section{Further Analysis of Motivation and Validity}
In this section, we provide additional analysis to support the motivations and validity of our design choices. We first examine the motivation and effectiveness of weighted cluster aggregation. We then analyze the rationale and empirical validity of token pruning.
\subsection{Weighted Aggregation}
\begin{figure}[h]
\centering
\includegraphics[width=\linewidth]{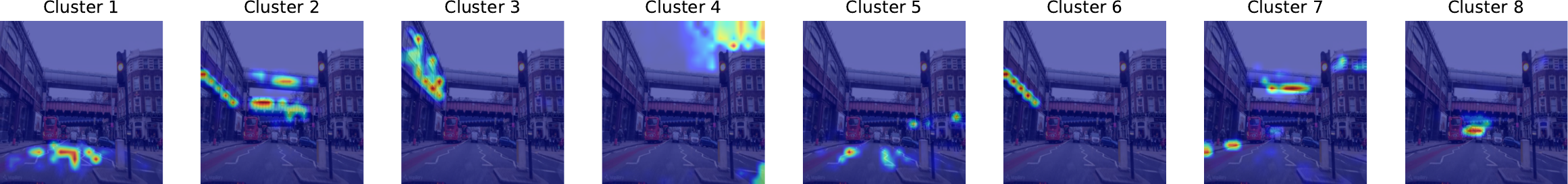}
\caption{Cluster-to-patch transport heatmaps showing that spatial and semantic patterns of different clusters are distinct.}
\label{fig:cluster_spasemdiff}
\end{figure}

As discussed in Section~\ref{sec:intro} of the main paper, clusters learned by VLAD-style aggregation are not homogeneous. They exhibit clear differences in both semantic content and spatial distribution. As illustrated in Fig.~\ref{fig:cluster_spasemdiff}, some clusters predominantly absorb patches corresponding to sky or background regions, which are typically less informative for VPR. In contrast, other clusters focus on structural elements such as buildings, which are more robust to viewpoint and scale variations~\cite{SelaVPR}. In addition, certain clusters demonstrate strong spatial preferences, activating consistently in specific image regions. These semantic and spatial disparities suggest that different clusters may contribute unequally to VPR performance. To empirically verify this observation, we conduct a controlled analysis within the SALAD framework: we construct global descriptors using tokens assigned to a single cluster $j$ at a time, \ie, $\mathbf{v}_j = \sum_{i=1}^{N} \mathbf{P}^\star_{ij}\,\tilde{\mathbf{x}}_i$, while discarding all others. On the MSLS-Val dataset, the resulting Recall@1 varies substantially across clusters, ranging from 34.6\% (worst-performing cluster) to 49.9\% (best-performing cluster), with a gap of 15.3\%.

This large performance variance provides direct evidence that clusters contribute to VPR with differing degrees of importance. We revisit a traditional method for recognition, \ie, vocabulary tree~\cite{ScalableReco}, which applies a weighting mechanism that suppresses nodes with lower discriminative power (such as common background textures) by assigning them lower weights, whilst assigning higher weights to nodes with greater discriminative power. Inspired by this, we ask whether these inherent differences can be explicitly exploited: 
\begin{tcolorbox}
Specifically, we ask whether amplifying clusters with higher VPR relevance while suppressing less informative ones can lead to a stronger global representation. 
\end{tcolorbox}
This question forms the basis of our weighted aggregation design.

Beyond quantitative analysis, Fig.~\ref{fig:atten} provides qualitative support for this motivation. Compared with uniform aggregation, the weighted global descriptor exhibits higher discriminability, with stronger responses on structurally stable regions such as buildings while attenuating background-dominated areas. This behavior is consistent with the cluster-level semantic and spatial biases discussed above. Quantitatively, the ablation study in Tab.~\ref{tab:aba} further confirms the effectiveness of weighted aggregation. Introducing cluster weighting consistently improves overall VPR performance across benchmarks, validating that explicitly modeling unequal cluster importance leads to a more informative global representation.

\subsection{Token Pruning for VPR}
Having established the necessity of weighted aggregation, we next investigate how such importance can be estimated early and exploited for efficient token pruning. Efficiency is a critical factor for the practical deployment of VPR systems. Even millisecond-level improvements in global descriptor extraction can be substantially amplified in large-scale settings, such as segment-level coarse retrieval, multi-stage retrieval pipelines, and city-scale applications involving millions of images (\eg, urban navigation~\cite{VisualNavigation}, UAV localization~\cite{VPAir}), as well as downstream tasks such as visual SLAM~\cite{VisualSLAM} and 3D reconstruction~\cite{3DRecon}.

Most existing efficiency-oriented VPR methods focus on descriptor-level compression to accelerate retrieval and reduce storage requirements. However, feature extraction typically dominates the overall computational budget and directly affects all subsequent stages. Improving feature extraction speed is commonly achieved by sacrificing representational capacity, for example by adopting lighter CNN-based backbones or smaller Vision Transformers. In contrast, as discussed in the previous subsection, many patch tokens correspond to regions that are uninformative or even detrimental to VPR performance, such as sky or road areas. 
\begin{tcolorbox}
This observation naturally raises the following question: can such tokens be discarded at an early stage to simultaneously improve efficiency and maintain or even enhance performance? 
\end{tcolorbox}
This question forms the basis of our token pruning design.

To the best of our knowledge, this work is the first to introduce token pruning into VPR. As qualitatively shown in Fig.~\ref{fig:tp}, the proposed pruning module is able to discard uninformative or low-importance patch tokens at early layers, while preserving tokens corresponding to structurally stable regions. Quantitatively, Tab.~\ref{tab:vpr_performance} and \ref{tab:Appendix_full_results} further validate the effectiveness of this strategy. At relatively high retention ratios, removing harmful tokens can even lead to improved VPR performance. When more tokens are pruned, performance degrades gracefully, while computational speed improves significantly. Importantly, this pruning mechanism is applied only at inference time and allows the retention ratio to be adjusted on the fly, enabling a flexible afficiency–eccuracy trade-off without retraining. This elasticity allows a single model to operate in either a high-accuracy or high-speed regime depending on downstream requirements, which underlies our \emph{faster or stronger} design motivation.

A natural question is why pruning is needed when smaller backbones (\eg, ViT-S) can already be faster and sometimes more accurate than a pruned larger model (\eg, ViT-B). The key distinction lies in flexibility. Choosing a larger backbone determines a higher performance upper bound, while token pruning provides a continuous and adaptive mechanism to trade accuracy for efficiency at inference time. These two approaches therefore play complementary roles rather than competing ones. As further demonstrated in Fig.~\ref{fig:diff_backbone} and Tab.~\ref{tab:diff_size}, the proposed pruning module generalizes well across Vision Transformers of different sizes, confirming its broad applicability. 

\section{Additional Algorithmic Details}
\subsection{Optimal Transport Formulation}
The soft assignment for aggregation is defined as an entropy-regularized OT problem. Following SuperGlue~\cite{SuperGlue} and SALAD~\cite{SALAD}, the classical Sinkhorn algorithm~\cite{SinkHorn, SinkHornOT} can be formulated as (left):
\begin{equation}
\min_{P\in\Pi(a,b)} \langle \mathbf{P}, \mathbf{C}\rangle + \varepsilon \sum_{i,j} \mathbf{P}_{ij}\log \mathbf{P}_{ij} \quad \rightarrow \quad
\max_{P\in\Pi(a,b)} \;\langle \mathbf{P}, \tilde{\mathbf{S}}\rangle + \varepsilon \sum_{i,j} \mathbf{P}_{ij}\log \mathbf{P}_{ij}.
\end{equation}
This variant formulation (right) is equivalent to the classical entropy-regularized OT problem with cost matrix $C$ by setting $S = -C$, but is more convenient for similarity-based matching, as it directly maximizes affinities rather than minimizing costs. The feasible set
\begin{equation}
\Pi(a,b)=\left\{  \mathbf{P}\ge 0 \;\middle|\;  \mathbf{P}\mathbf{1}=a,\;  \mathbf{P}^\top\mathbf{1}=b \right\} , \quad P\in\mathbb{R}_+^{(N+1)\times (M+1)}, \, a\in\mathbb{R}^{N+1}_+, \, b\in\mathbb{R}^{M+1}_+
\end{equation}
enforces marginal constraints with source mass $a$ and target mass $b$. The additional row and column correspond to a ghost token and a dustbin cluster, respectively, enabling the model to absorb uninformative or unmatched content in a bidirectional manner.

\subsection{Auxiliary Projection and Prediction Modules}
In this subsection, we provide additional details of the lightweight modules used for dimension reduction, transport score projection, and early importance prediction. All modules are implemented as shallow MLPs and introduce negligible computational overhead. Dimension reduction and score projection are adopted from SALAD~\cite{SALAD}.

\paragraph{Dimension Reduction}

We apply two lightweight MLPs to project patch tokens and the \texttt{[CLS]} token into lower-dimensional spaces before OT matching.
Given patch token embeddings $\mathbf{X} = [\mathbf{x}_1, \ldots, \mathbf{x}_N] \in \mathbb{R}^{N \times D}$ and the \texttt{[CLS]} token $\mathbf{x}_{\texttt{CLS}} \in \mathbb{R}^{D}$, the projections are defined as:
\begin{equation}
    \tilde{\mathbf{X}} = f_1(\mathbf{X})
    = W^{(1)}_2 \, \sigma\!\left(W^{(1)}_1 \mathbf{X} + \mathbf{b}^{(1)}_1\right) + \mathbf{b}^{(1)}_2,
\end{equation}
\begin{equation}
    \tilde{\mathbf{x}}_{\texttt{CLS}} = f_2(\mathbf{x}_{\texttt{CLS}})
    = W^{(2)}_2 \, \sigma\!\left(W^{(2)}_1 \mathbf{x}_{\texttt{CLS}} + \mathbf{b}^{(2)}_1\right) + \mathbf{b}^{(2)}_2,
\end{equation}
where $\sigma(\cdot)$ denotes the ReLU activation. Both $f_1$ and $f_2$ reduce the feature dimensionality from $D$ to $d \ll D$, which stabilizes OT matching and reduces computational cost.

\paragraph{Score Projection}
To compute affinities between patch tokens and clusters, we learn a transport score matrix using another lightweight MLP:
\begin{equation}
    \mathbf{S} = f_s(\tilde{\mathbf{X}})
    = W^{(s)}_2 \, \sigma\!\left(W^{(s)}_1 \tilde{\mathbf{X}} + \mathbf{b}^{(s)}_1\right) + \mathbf{b}^{(s)}_2,
\end{equation}
where $\mathbf{S} \in \mathbb{R}^{N \times M}$ and $\mathbf{S}_{ij}$ measures the affinity between patch token $i$ and cluster $j$.
This score matrix serves as the similarity input to the entropy-regularized OT formulation described above.

\paragraph{Student Importance Predictor (WeiToP)}
To enable early token pruning, we introduce a lightweight student module, termed \textbf{WeiToP}, which predicts token importance from early-layer features. Specifically, WeiToP operates on token embeddings $\mathbf{X}^{(s)} \in \mathbb{R}^{N_0 \times D}$ extracted from the first transformer block and outputs importance logits:
\begin{equation}
    \mathbf{z}^{(s)} = f_z(\mathbf{X}^{(s)})
    = W^{(z)}_2 \, \sigma\!\left(W^{(z)}_1 \mathbf{X}^{(s)} + \mathbf{b}^{(z)}_1\right) + \mathbf{b}^{(z)}_2,
\end{equation}
where $\mathbf{z}^{(s)} = \{z^{(s)}_i\}_{i=1}^{N_0}$. During training, WeiToP is supervised by aggregation-derived token importance scores $\mathcal{I}_i$ through a distillation objective, while at inference time it enables adaptive token pruning without modifying the backbone.




\section{Dataset Description}\label{sec:datadescri}
\begin{table*}[h]
  \centering
  \resizebox{\textwidth}{!}{
\begin{tabular}{ccccccc}
\hline
\textbf{Dataset}     & \textbf{MSLS-C} & \textbf{MSLS-val} & \textbf{Nordland} & \textbf{Pitts250k-test} & \textbf{SPED} & \textbf{AmsterTime} \\ \hline
\textbf{Num. of Q.}  & 27092   & 740      & 2706     & 8280    & 607 & 1231 \\
\textbf{Num. of R.}  & 38770   & 18871    & 27592    & 83952   & 607 & 1231 \\ \hdashline
\textbf{Characteristic} &  \begin{tabular}[c]{@{}c@{}}Long-term\\Online challenge\\ Close evaluation\end{tabular} &  \begin{tabular}[c]{@{}c@{}}Temporal diversity\\ Large-scale\\ Seasonal variations\end{tabular} & \begin{tabular}[c]{@{}c@{}}Fixed viewpoints\\ Temporal shifts\\ Seasonal variation\end{tabular} &  \begin{tabular}[c]{@{}c@{}}Visual distractors\\ Large-scale\\ Varying poses\end{tabular} & \begin{tabular}[c]{@{}c@{}}Fixed viewpoints\\ Lighting variations\\ Environmental challenges\end{tabular} & \begin{tabular}[c]{@{}c@{}}Temporal shifts\\ Modality challenge\\ Viewpoint variations\end{tabular} \\ \hline
\end{tabular}
  }
  \caption{Dataset details for comparison.}
  \label{tab:dtdetails2}
\end{table*}

\textbf{GSV-Cities} dataset~\cite{GSV-Cities} is a large-scale, high-precision dataset covering locations worldwide. Owing to its accurate annotations and rich diversity, it has been widely adopted as a high-quality training source for modern deep learning models.

\textbf{MSLS} dataset~\cite{MSLS} is also a large-scale dataset covering 30 cities, specifically created for image- and sequence-based visual place recognition (VPR). It contains over one million images from multiple cities, although only a small subset is used for evaluation. Following common practice~\cite{EigenPlaces,CosPlace}, we evaluate on its validation set. The dataset incorporates diverse viewpoints, long-term seasonal changes, and dynamic urban objects, and also includes a small number of nighttime and side-view images. In addition, a more stringent online benchmark is provided for evaluation and comparison (hosted on the CodaLab online platform\footnote{Since the CodaLab platform officially stopped accepting submissions on January 20, submissions to the MSLS challenge have also been suspended.}).

\textbf{Pitts} dataset~\cite{NetVLAD} is one of the most commonly used benchmarks in VPR. It is constructed using Google Street View images from downtown Pittsburgh, ensuring that the database and query images are collected from different years. It is mainly used to evaluate the robustness of algorithms to significant viewpoint changes in urban environments. We use the test set of the larger Pitts250k split for evaluation.

\textbf{Nordland} dataset~\cite{Nordland} consists of perfectly aligned railway videos captured across four seasons, and serves as a standard benchmark for testing pure appearance robustness to severe seasonal changes without considering viewpoint variations. Similar to previous works~\cite{SALAD}, we use the summer trajectory as the query set and all remaining data as the reference set.

\textbf{SPED} dataset~\cite{SPED} is composed of images captured by surveillance cameras around the world. We use a subset of this dataset, which contains images captured by fixed outdoor surveillance cameras and covers extreme appearance changes such as day–night transitions, seasonal variations, and weather changes. This dataset emphasizes long-term visual changes under fixed viewpoints, making it particularly suitable for evaluating robustness to illumination and temporal variations in place recognition tasks.

\textbf{AmsterTime} dataset~\cite{AmsterTime} combines historical archival images with modern street-view imagery, challenging algorithms to recognize places across long time spans and structural environmental changes. These image pairs exhibit multiple domain gaps, including viewpoint differences, long-term temporal changes, modality differences (RGB vs. grayscale), and different camera types. As a result, AmsterTime is one of the most challenging datasets.

\section{Implementation Details of Comparative Methods}
\subsection{VPR Methods}\label{sec:implemencompare}
\textbf{NetVLAD}~\cite{NetVLAD} is the pioneer of deep learning in visual localization, transforming traditional VLAD algorithms into differentiable neural network layers to achieve end-to-end feature aggregation.

\textbf{CosPlace}~\cite{CosPlace} abandons complex metric learning. It reframes location recognition as a large-scale classification problem, demonstrating that simple aggregation paired with a strong classification loss function can also achieve SOTA performance.

\textbf{Conv-AP}~\cite{GSV-Cities} pioneers a novel approach by directly optimizing the final retrieval metric Average Precision (AP) during training, thereby maximizing the ranking quality of retrieval results.

\textbf{MixVPR}~\cite{MixVPR} is a fast and lightweight model, employing a fully MLP architecture to fuse features across channels and spatial dimensions, achieving inference efficiency far surpassing traditional convolutional models.

\textbf{EigenPlaces}~\cite{EigenPlaces} focuses on the challenge of extreme viewpoint variations by establishing eigenclasses through mining images from different viewpoints of the same location, achieving exceptional viewpoint robustness.

\textbf{SuperVLAD}\footnote{https://github.com/lu-feng/SuperVLAD}~\cite{SuperVLAD} is a compact refinement of NetVLAD, aiming to generate lower-dimensional, more generalizable global descriptors through decentralized or optimized codebooks.

\textbf{SALAD}\footnote{https://github.com/serizba/salad}~\cite{SALAD} incorporates optimal transport theory and a “garbage can” mechanism to efficiently filter background noise like sky and moving objects, precisely aggregating landmark features.

For NetVLAD, Conv-AP, MixVPR, and EigenPlaces, we adopt the results reported in SALAD, as their experimental settings are consistent with ours. For SuperVLAD, SALAD, and our methods (WeiAD and WeiAD+WeiToP), we measure inference latency on a single NVIDIA RTX 3090 GPU, using the same evaluation settings as SALAD, following the evaluation protocol of R2Former\footnote{https://github.com/bytedance/R2Former}, using an input resolution of 322×322 and a batch size of 256. SuperVLAD is evaluated using the official DINOv2 ViT-B checkpoint with the additional cross-image encoder disabled. Since our method is built upon SALAD, we report results for both SALAD and our methods averaged over five runs with different random seeds.

\subsection{Token Pruning Methods}\label{sec:implemencompare2}
\textbf{DynamicViT}\footnote{https://github.com/raoyongming/DynamicViT}~\cite{DynamicViT} dynamically generates masks by incorporating learnable prediction modules between layers, enabling it to directly discard unimportant background tokens. We integrate DynamicViT into WeiAD by embedding its three-stage learnable token prediction modules into the backbone network. Models are trained with different token retention ratios at each stage (0.86, 0.80, 0.75, 0.70, 0.60, 0.50, 0.47), resulting in overall retention ratios $\rho$ of (0.64, 0.51, 0.42, 0.34, 0.22, 0.13, 0.10). All other settings follow the original paper, and token pruning is applied after the 3rd, 6th, and 9th layers.

\textbf{ToMe}\footnote{https://github.com/facebookresearch/ToMe}~\cite{ToMe} employs a binary matching algorithm to pair and merge similar tokens, achieving lossless reduction without retraining. We integrate the token matching and merging mechanism of ToMe into WeiAD. Following the original paper, and for better performance, ToMe is re-trained with different numbers of tokens merged at each layer (17, 19, 25, 30, 40, 50), corresponding to overall retention ratios $\rho$ of (0.62, 0.58, 0.44, 0.33, 0.11, 0.02). All other settings follow the original paper, and token merging is applied after each transformer layer.

\textbf{ToFu}~\cite{ToFu} absorbs redundant token information into core tokens through clustering or feature aggregation, achieving higher fidelity than simple pruning. We integrate ToFu into WeiAD by adopting its token clustering and feature aggregation strategy as the token reduction module. Models are trained with different per-layer token retention ratios $\rho$ (0.96, 0.94, 0.92, 0.90, 0.85, 0.80), corresponding to overall keep ratios of (0.61, 0.48, 0.37, 0.28, 0.14, 0.07). All other settings follow the original paper. A hybrid token reduction strategy is employed, where token pruning is applied after each of the first 6 layers, and token merging is applied after each of the last 6 layers.

\textbf{FastV}\footnote{https://github.com/pkunlp-icler/FastV}~\cite{FastV} targets multimodal large models by directly eliminating low-contribution visual tokens in deep networks based on attention weight rankings. In contrast, FastV is applied to WeiAD in a training-free manner. We directly adopt its attention-based token importance scores to prune visual tokens during the inference stage of WeiAD, with retention ratios $\rho$ of 0.95, 0.70, 0.50, and 0.40. All other settings follow the original paper, and pruning is applied after the 2nd layer.

All methods are trained or evaluated five times with different random seeds for each configuration.

\section{Complete Result}
\subsection{General Result}\label{sec:completeresults}

\begin{table*}[t]
\caption{Complete results of \textbf{WeiAD} + \textbf{WeiToP}, our reproduced \textbf{SALAD}, and \textbf{WeiAD} + \textbf{DynamicViT}, + \textbf{ToMe}, + \textbf{ToFu}, and + \textbf{FastV} token pruning strategies. The best results without pruning are highlighted in \textbf{bold}. \redbf{Bold red} indicates that the performance after pruning still achieves or surpasses the best results. *The MSLS challenge suspended submissions due to the closure of platform on Jan 20th in 2026, so we can only report partial results.}\label{tab:Appendix_full_results}
\resizebox{\linewidth}{!}{
\begin{tabular}{lcccccccccccccc>{\columncolor{gray!10}}c>{\columncolor{gray!10}}c}
\toprule[1.5pt]
\multirow{2}{*}{\textbf{Method}} & \multirow{2}{*}{$\boldsymbol{\rho}$} & \textbf{Latency} & \multicolumn{2}{c}{\textbf{MSLS-C}}         & \multicolumn{2}{c}{\textbf{MSLS-Val}}    & \multicolumn{2}{c}{\textbf{NordLand}}    & \multicolumn{2}{c}{\textbf{Pitts250k-test}} & \multicolumn{2}{c}{\textbf{SPED}}        & \multicolumn{2}{c}{\textbf{AmsterTime}} & \multicolumn{2}{>{\columncolor{gray!10}}c}{\textbf{Mean} \text{\scriptsize w/o MSLS-C}} \\ \cline{4-17} 
                        &                          &            (ms)               & R@1            & R@5               & R@1            & R@5            & R@1           & R@5             & R@1            & R@5               & R@1            & R@5            & R@1           & R@5          & R@1           & R@5           \\ \midrule[1.5pt]
                        
\rowcolor{gray!15}\multicolumn{17}{l}{\textit{VPR Methods}} \\
SALAD                   & /                        & \ms{2.41}{0.00}                          & 
 \ms{75.4}{0.70} &  \ms{87.4}{0.61}  &
 \ms{91.4}{0.69} & \ms{\textbf{96.2}}{0.28} &
 \ms{71.1}{2.45} & \ms{85.6}{2.03} &
 \ms{94.9}{0.20} & \ms{98.5}{0.20} &
 \ms{90.3}{0.50} & \ms{\textbf{95.4}}{0.36} &
 \ms{58.0}{1.10} & \ms{78.6}{0.98} & \ms{81.1}{0.99} & \ms{90.9}{0.77} \\ 
WeiAD              & 100                      &  \ms{2.41}{0.00}                           &
 \ms{\textbf{76.2}}{0.32} &  \ms{\textbf{87.8}}{0.54}  &
 \ms{\textbf{91.7}}{0.47} & \ms{\textbf{96.2}}{0.26} &
 \ms{\textbf{73.1}}{2.43} & \ms{\textbf{86.7}}{2.02} &
 \ms{\textbf{95.1}}{0.20} & \ms{\textbf{98.8}}{0.07} &
 \ms{\textbf{90.6}}{0.30} & \ms{\textbf{95.4}}{0.44} &
 \ms{\textbf{58.4}}{0.72} & \ms{\textbf{79.5}}{0.75} &\ms{\textbf{81.8}}{0.82} & \ms{\textbf{91.3}}{0.71} \\ \hline
\rowcolor{gray!15}\multicolumn{17}{l}{\textit{WeiAD (+ Token Pruning)}} \\ \hdashline
\quad \multirow{9}{*}{\rot{+ WeiToP}} & 0.95                     & \ms{2.23}{0.00}    &                      \ms{\redbf{76.3}}{0.43} &  \ms{\redbf{87.9}}{0.58}  & \ms{\redbf{91.8}}{0.60} & \ms{\redbf{96.3}}{0.10} & \ms{72.8}{2.22} & \ms{\redbf{86.7}}{1.81} & \ms{95.0}{0.23} & \ms{98.6}{0.05} & \ms{90.4}{0.18} & \ms{95.3}{0.42} & \ms{\redbf{58.4}}{0.88} & \ms{79.4}{0.76} & \ms{81.7}{0.82} & \ms{91.2}{0.63} \\
           & 0.90                     & \ms{2.13}{0.00}                           &
   -        &          -         & \ms{\redbf{91.8}}{0.68} & \ms{\redbf{96.3}}{0.17} &
 \ms{71.5}{2.21} & \ms{85.9}{1.97} &
 \ms{94.9}{0.25} & \ms{98.6}{0.03} &
 \ms{90.0}{0.16} & \ms{95.2}{0.34} &
 \ms{57.7}{1.00} & \ms{79.1}{0.70} & \ms{81.2}{0.86} & \ms{91.0}{0.64} \\ 
            & 0.85                     & \ms{2.03}{0.01}                           &     -           &         -          & \ms{\redbf{91.9}}{0.64} & \ms{\redbf{96.3}}{0.17} &
 \ms{70.8}{2.20} & \ms{85.5}{2.06} &
 \ms{94.8}{0.22} & \ms{98.6}{0.04} &
 \ms{89.6}{0.28} & \ms{95.0}{0.28} &
 \ms{56.9}{0.93} & \ms{78.6}{0.87} & \ms{80.8}{0.85} & \ms{90.8}{0.68} \\
              & 0.80                     & \ms{1.94}{0.01}                           &       -         &     -              & \ms{\redbf{91.8}}{0.46} & \ms{\redbf{96.3}}{0.22} &
 \ms{70.5}{2.03} & \ms{85.1}{2.08} &
 \ms{94.6}{0.25} & \ms{98.5}{0.06} &
 \ms{89.6}{0.54} & \ms{94.7}{0.46} &
 \ms{56.6}{1.24} & \ms{78.2}{0.67} & \ms{80.6}{0.90} & \ms{90.6}{0.70} \\
             & 0.75                     & \ms{1.83}{0.00}                           &         -       &     -              & \ms{91.6}{0.42} & \ms{96.1}{0.14} &
 \ms{70.1}{1.69} & \ms{84.8}{2.08} &
 \ms{94.3}{0.22} & \ms{98.4}{0.05} &
 \ms{89.5}{0.64} & \ms{94.6}{0.45} &
 \ms{56.0}{1.28} & \ms{77.8}{1.04} & \ms{80.3}{0.85} & \ms{90.3}{0.75} \\ 
             & 0.70                     & \ms{1.68}{0.01}                          & 
\ms{75.3}{0.34} &  \ms{86.8}{0.58}  & \ms{91.3}{0.31} & \ms{96.1}{0.34} &
 \ms{69.5}{1.42} & \ms{84.7}{2.03} &
 \ms{94.0}{0.16} & \ms{98.3}{0.06} &
 \ms{89.2}{0.72} & \ms{94.4}{0.36} &
 \ms{55.3}{1.06} & \ms{76.9}{0.77} & \ms{79.9}{0.73} & \ms{90.1}{0.71} \\ 
            & 0.60                     & \ms{1.50}{0.01}                          &        -        &      -             & \ms{91.1}{0.61} & \ms{96.0}{0.27} &
 \ms{67.6}{1.49} & \ms{83.6}{1.64} &
 \ms{93.1}{0.23} & \ms{98.0}{0.09} &
 \ms{87.5}{0.43} & \ms{93.8}{0.30} &
 \ms{53.0}{1.08} & \ms{75.0}{0.63} & \ms{78.5}{0.77} & \ms{89.3}{0.58} \\ 
             & 0.50                     & \ms{1.30}{0.00}                           &  
    \ms{71.6}{0.54} &  \ms{84.2}{0.42}  & \ms{90.4}{0.73} & \ms{95.9}{0.30} &
 \ms{62.3}{1.43} & \ms{80.0}{1.52} &
 \ms{91.5}{0.39} & \ms{97.4}{0.12} &
 \ms{85.5}{0.44} & \ms{92.5}{0.40} &
 \ms{48.7}{1.20} & \ms{70.6}{1.67} & \ms{75.7}{0.84} & \ms{87.3}{0.80} \\ 
             & 0.40                     & \ms{1.09}{0.00}                           &
\ms{68.2}{0.64} &  \ms{81.9}{0.71}  &\ms{87.7}{0.88} & \ms{94.8}{0.47} &
 \ms{53.3}{1.70} & \ms{71.0}{1.55} &
 \ms{88.2}{0.50} & \ms{96.2}{0.22} &
 \ms{82.4}{1.22} & \ms{90.9}{0.60} &
 \ms{41.8}{1.71} & \ms{64.0}{1.68} & \ms{70.7}{1.20} & \ms{83.4}{0.90} \\ \hdashline 
\quad \multirow{7}{*}{\rot{+ DynamicViT}}  & 0.64                     & \ms{2.05}{0.01} & - & - &
 \ms{87.6}{2.55} & \ms{94.5}{0.87} &
 \ms{56.0}{2.89} & \ms{73.9}{3.08} &
 \ms{93.1}{0.43} & \ms{97.6}{0.42} &
 \ms{88.0}{0.95} & \ms{93.7}{0.81} &
 \ms{50.4}{2.92} & \ms{72.1}{3.07} & \ms{75.0}{1.95} & \ms{86.4}{1.65} \\
      & 0.51                     & \ms{1.87}{0.01} & - & - &
 \ms{86.0}{1.84} & \ms{93.1}{1.13} &
 \ms{45.6}{9.42} & \ms{65.0}{9.72} &
 \ms{92.5}{0.46} & \ms{97.4}{0.38} &
 \ms{85.9}{1.42} & \ms{93.4}{0.53} &
 \ms{48.6}{3.04} & \ms{70.7}{2.67} & \ms{71.7}{3.24} & \ms{83.9}{2.89} \\ 
     & 0.42                     & \ms{1.75}{0.00} & - & - &
 \ms{84.5}{3.21} & \ms{92.0}{2.14} &
 \ms{42.8}{11.22} & \ms{61.5}{12.32} &
 \ms{92.1}{0.51} & \ms{97.2}{0.32} &
 \ms{84.4}{2.37} & \ms{93.2}{1.03} &
 \ms{47.1}{2.81} & \ms{69.4}{3.72} & \ms{70.2}{4.03} & \ms{82.7}{3.91} \\
      & 0.34                     & \ms{1.60}{0.00} & - & - &
 \ms{78.2}{4.52} & \ms{88.6}{2.66} &
 \ms{34.0}{15.20} & \ms{51.7}{17.80} &
 \ms{90.1}{1.14} & \ms{96.2}{0.71} &
 \ms{82.1}{2.90} & \ms{91.4}{1.25} &
 \ms{39.3}{2.95} & \ms{61.6}{3.29} & \ms{64.7}{5.34} & \ms{77.9}{5.14} \\ 
       & 0.22                     & \ms{1.39}{0.00} & - & - &
 \ms{71.7}{6.44} & \ms{84.9}{4.03} &
 \ms{20.9}{9.16} & \ms{35.7}{13.22} &
 \ms{86.3}{2.12} & \ms{94.8}{1.27} &
 \ms{72.6}{7.50} & \ms{84.7}{5.96} &
 \ms{33.7}{1.32} & \ms{55.0}{2.98} & \ms{57.0}{5.31} & \ms{71.1}{5.49} \\
        & 0.13                     & \ms{1.20}{0.01} & - & - &
 \ms{60.9}{8.70} & \ms{77.2}{7.06} &
 \ms{12.4}{7.24} & \ms{24.1}{12.27} &
 \ms{79.2}{2.80} & \ms{91.9}{1.49} &
 \ms{63.2}{6.01} & \ms{78.8}{3.82} &
 \ms{25.9}{5.70} & \ms{45.8}{7.53} & \ms{48.3}{6.09} & \ms{63.5}{6.43} \\
       & 0.10                     & \ms{1.15}{0.01} & - & - &
 \ms{60.8}{6.11} & \ms{77.3}{4.71} &
 \ms{11.8}{5.71} & \ms{23.8}{9.84} &
 \ms{74.5}{2.23} & \ms{88.3}{1.54} &
 \ms{55.6}{7.49} & \ms{72.2}{6.11} &
 \ms{27.4}{3.88} & \ms{48.0}{4.37} & \ms{46.0}{5.08} & \ms{61.9}{5.31} \\ \hdashline 
\quad \multirow{6}{*}{\rot{+ ToMe}}    & 0.62                     & \ms{2.15}{0.00} & - & - &
 \ms{89.8}{0.45} & \ms{95.2}{0.35} &
 \ms{64.4}{2.08} & \ms{81.4}{1.38} &
 \ms{93.8}{0.27} & \ms{98.1}{0.11} &
 \ms{88.6}{0.58} & \ms{94.9}{0.41} &
 \ms{52.6}{1.14} & \ms{74.9}{0.78} & \ms{77.8}{0.91} & \ms{88.9}{0.61} \\
         & 0.58                     & \ms{2.09}{0.00}& - & - &
 \ms{88.1}{0.77} & \ms{94.6}{0.72} &
 \ms{64.6}{2.79} & \ms{82.4}{2.06} &
 \ms{93.4}{0.30} & \ms{97.9}{0.17} &
 \ms{86.7}{0.28} & \ms{93.8}{0.89} &
 \ms{50.5}{0.47} & \ms{73.1}{0.82} & \ms{76.7}{0.92} & \ms{88.4}{0.93} \\
         & 0.44                     & \ms{1.91}{0.01} & - & - &
 \ms{80.7}{0.79} & \ms{89.4}{0.32} &
 \ms{29.1}{2.50} & \ms{48.2}{2.98} &
 \ms{87.5}{0.41} & \ms{95.4}{0.27} &
 \ms{77.4}{0.56} & \ms{88.5}{0.93} &
 \ms{39.4}{0.82} & \ms{62.3}{1.54} & \ms{62.8}{1.02} & \ms{76.7}{1.21} \\
         & 0.33                     & \ms{1.78}{0.01} & - & - &
 \ms{71.1}{1.99} & \ms{83.6}{1.35} &
 \ms{12.6}{0.75} & \ms{23.0}{1.72} &
 \ms{80.8}{1.25} & \ms{91.4}{0.63} &
 \ms{69.4}{1.86} & \ms{82.6}{1.68} &
 \ms{32.8}{2.20} & \ms{54.9}{3.06} & \ms{53.3}{1.61} & \ms{67.1}{1.69} \\
          & 0.11                     & \ms{1.50}{0.01} & - & - &
 \ms{27.4}{4.02} & \ms{43.8}{6.32} &
 \ms{1.7}{0.38} & \ms{4.0}{0.72} &
 \ms{44.2}{2.90} & \ms{67.4}{3.58} &
 \ms{46.1}{3.83} & \ms{66.4}{3.24} &
 \ms{13.8}{1.31} & \ms{29.1}{3.53} & \ms{26.6}{2.49} & \ms{42.1}{3.48} \\ 
          & 0.02                     & \ms{1.26}{0.00} & - & - &
 \ms{10.4}{1.04} & \ms{21.0}{1.91} &
 \ms{0.4}{0.04} & \ms{1.1}{0.20} &
 \ms{24.9}{1.35} & \ms{44.7}{1.65} &
 \ms{25.8}{2.80} & \ms{42.5}{3.32} &
 \ms{6.0}{0.41} & \ms{14.3}{1.04} & \ms{13.5}{1.13} & \ms{24.7}{1.62} \\ \hdashline 
\quad \multirow{6}{*}{\rot{+ ToFu}}  & 0.61                     & \ms{2.20}{0.00} & - & - &
 \ms{91.5}{0.80} & \ms{\redbf{96.3}}{0.17} &
 \ms{68.2}{2.53} & \ms{82.9}{2.17} &
 \ms{94.9}{0.27} & \ms{98.7}{0.13} &
 \ms{90.3}{0.51} & \ms{\redbf{95.7}}{0.34} &
 \ms{56.6}{0.14} & \ms{78.2}{0.80} & \ms{80.3}{0.85} & \ms{90.4}{0.72} \\
         & 0.48                     & \ms{2.00}{0.01} & - & - &
 \ms{91.3}{0.64} & \ms{\redbf{96.2}}{0.15} &
 \ms{69.0}{1.98} & \ms{84.1}{1.97} &
 \ms{94.3}{0.11} & \ms{98.5}{0.26} &
 \ms{89.5}{0.74} & \ms{95.4}{0.37} &
 \ms{55.7}{0.57} & \ms{77.2}{0.74} & \ms{80.0}{0.81} & \ms{89.3}{0.70} \\
         & 0.37                     & \ms{1.80}{0.00} & - & - &
 \ms{89.3}{0.48} & \ms{95.5}{0.07} &
 \ms{66.5}{1.62} & \ms{83.5}{1.17} &
 \ms{93.6}{0.15} & \ms{98.2}{0.10} &
 \ms{90.0}{0.82} & \ms{94.9}{0.62} &
 \ms{55.9}{0.60} & \ms{76.0}{1.04} & \ms{79.0}{0.73} & \ms{89.6}{0.60} \\
         & 0.28                     & \ms{1.68}{0.01} & - & - &
 \ms{89.7}{0.61} & \ms{95.7}{0.57} &
 \ms{62.6}{1.02} & \ms{78.8}{0.98} &
 \ms{93.7}{0.21} & \ms{98.1}{0.13} &
 \ms{88.1}{1.07} & \ms{94.6}{0.23} &
 \ms{51.1}{0.77} & \ms{73.5}{1.03} & \ms{77.0}{0.74} & \ms{88.1}{0.59} \\ 
        & 0.14                     & \ms{1.34}{0.00} & - & - &
 \ms{86.9}{2.21} & \ms{95.0}{1.60} &
 \ms{54.0}{0.54} & \ms{72.4}{0.81} &
 \ms{89.7}{1.04} & \ms{96.6}{0.65} &
 \ms{81.7}{1.01} & \ms{92.1}{0.51} &
 \ms{39.5}{0.81} & \ms{61.9}{0.96} & \ms{70.4}{1.12} & \ms{83.6}{0.91} \\  
        & 0.07                     & \ms{1.15}{0.00} & - & - &
 \ms{83.0}{0.45} & \ms{90.7}{1.41} &
 \ms{34.6}{0.05} & \ms{53.8}{0.12} &
 \ms{89.0}{0.47} & \ms{96.2}{1.20} &
 \ms{80.7}{2.05} & \ms{91.8}{1.75} &
 \ms{37.7}{0.26} & \ms{60.8}{0.37} & \ms{65.0}{0.66} & \ms{78.7}{0.97} \\  
\hdashline 
\quad \multirow{4}{*}{\rot{+ FastV}}             & 0.95                     & \ms{2.30}{0.00} & - & - &
 \ms{91.4}{0.11} & \ms{\redbf{96.3}}{0.23} &
 \ms{71.2}{2.14} & \ms{85.8}{1.93} &
 \ms{94.9}{0.27} & \ms{98.6}{0.19} &
 \ms{90.4}{0.16} & \ms{\redbf{95.4}}{0.21} &
 \ms{56.9}{0.68} & \ms{78.7}{0.67} & \ms{80.9}{0.67} & \ms{91.0}{0.65} \\ 
         & 0.70                     & \ms{1.86}{0.00} & - & - &
 \ms{89.5}{0.66} & \ms{95.3}{0.17} &
 \ms{60.9}{2.47} & \ms{78.7}{2.10} &
 \ms{93.8}{0.25} & \ms{98.4}{0.21} &
 \ms{89.1}{0.28} & \ms{95.2}{0.08} &
 \ms{49.4}{0.10} & \ms{71.0}{1.01} & \ms{76.5}{0.75} & \ms{87.7}{0.71} \\
       & 0.50                     & \ms{1.53}{0.00} & - & - &
 \ms{85.6}{0.07} & \ms{93.8}{0.62} &
 \ms{46.6}{1.69} & \ms{66.8}{1.92} &
 \ms{91.7}{0.25} & \ms{97.8}{0.15} &
 \ms{86.7}{0.64} & \ms{93.5}{0.28} &
 \ms{38.8}{0.34} & \ms{60.5}{0.84} & \ms{69.9}{0.60} & \ms{82.5}{0.76} \\  
       & 0.40                     & \ms{1.35}{0.00} & - & - &
 \ms{81.8}{1.44} & \ms{91.8}{0.77} &
 \ms{35.1}{1.17} & \ms{55.4}{1.58} &
 \ms{89.8}{0.32} & \ms{97.1}{0.13} &
 \ms{83.5}{0.61} & \ms{91.7}{0.64} &
 \ms{33.5}{0.87} & \ms{53.3}{0.91} & \ms{64.7}{0.88} & \ms{77.9}{0.81} \\
\bottomrule[1.5pt]
\end{tabular}
}
\end{table*}

\begin{figure}[t]
\centering
\includegraphics[width=\linewidth]{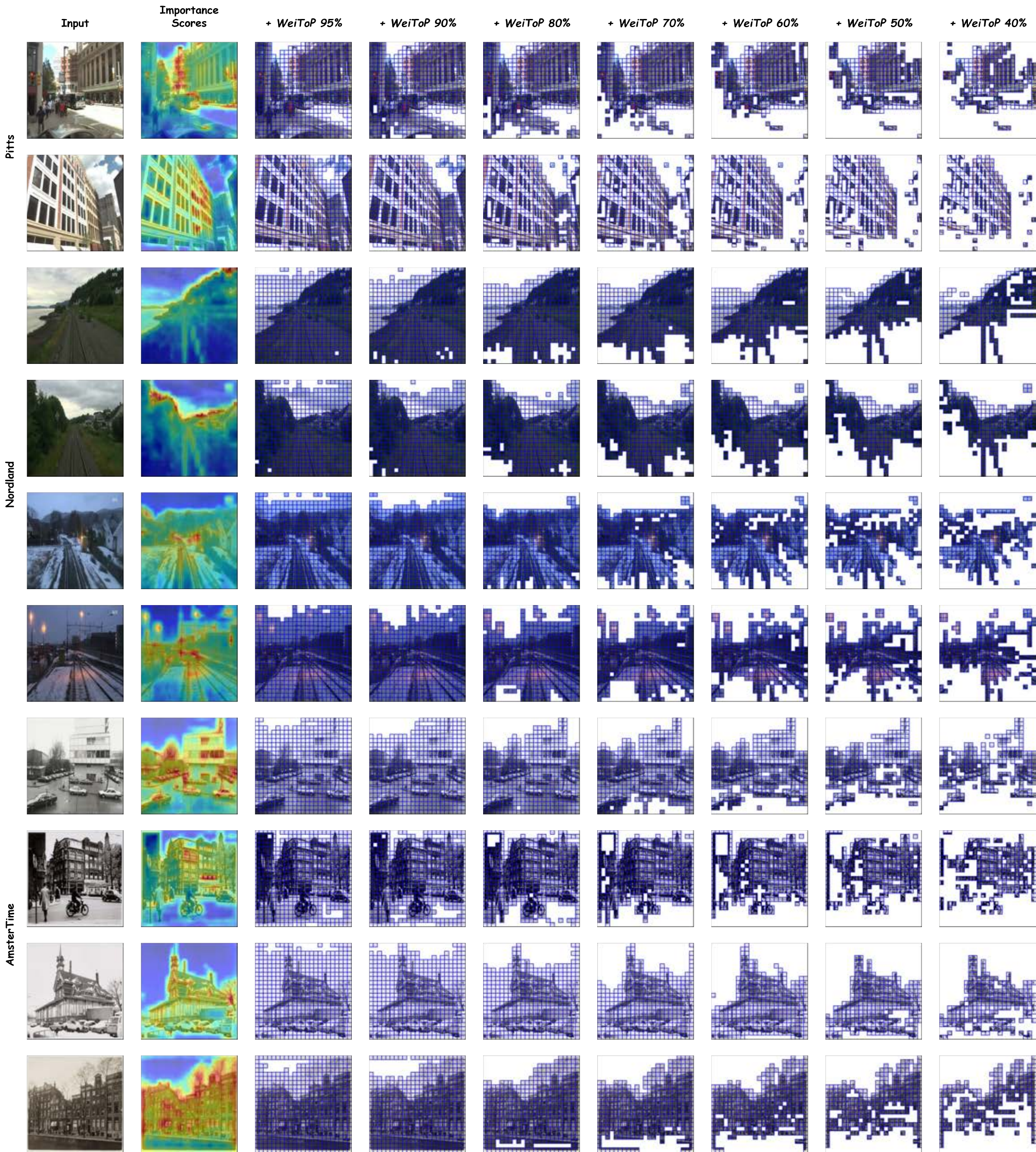}
\caption{Visual examples across different cities under different conditions. Heatmap of token importance scores and tokens retained after applying \textbf{WeiToP} with retention ratios $\rho=0.95$, $0.9$, $0.8$, $0.7$, $0.6$, $0.5$, and $0.4$. *\textcolor{gray}{Blank} indicates removed tokens.}
\label{fig:tp}
\end{figure}

Tab.\ref{tab:Appendix_full_results} shows the complete results from extensive experiments for comparison. Beyond the results discussed in Sec.~\ref{sec:Exp}, we observe that as $\rho$ is gradually reduced to 0.80, WeiToP consistently achieves the best or near-best performance on MSLS-Val, and remains close to the best-performing methods on Pitts250k-test. In contrast, a more noticeable performance degradation is observed on other datasets. These results suggest that for conventional urban street-view scenarios, WeiToP serves as an effective and flexible VPR solution, as redundant background regions can be safely discarded under standard viewpoints. However, for more challenging datasets, non-canonical viewpoints often contain harder-to-discriminate visual cues or exhibit little spatial redundancy across the entire image, making aggressive token pruning less favorable. The visual examples across different datasets in Fig.~\ref{fig:tp} further support this observation.

In addition, ToFu and FastV also achieve competitive or near-optimal performance on MSLS-Val and Pitts250k when the token retention ratio is relatively high. However, on more challenging datasets such as Nordland and AmsterTime, their performance drops more significantly compared to WeiToP. This indicates that although these methods can effectively remove redundant information under standard urban viewpoints, they are less robust when non-canonical viewpoints or globally informative scenes leave limited room for token redundancy. As one of the earliest baselines, DynamicViT still demonstrates reasonably strong performance on VPR tasks, with a moderate accuracy–efficiency trade-off. ToMe, which relies solely on token merging, exhibits less severe degradation at high keep ratios, but suffers from substantial performance loss when more aggressive reduction is required to further accelerate inference. In contrast, ToFu, which adopts a hybrid strategy combining token pruning and merging, better exploits the complementary strengths of both operations, achieving the strongest performance among existing methods aside from WeiToP and showing higher suitability for VPR. FastV, despite being training-free or requiring only lightweight joint learning, also achieves a favorable trade-off between efficiency and accuracy. Nevertheless, we observe that layer-wise token reduction methods such as DynamicViT, ToMe, and ToFu often require aggressive pruning (\ie, very low keep ratios) to reach lower latency, which may partially explain their abrupt performance degradation. Similarly, FastV applies pruning immediately after the second transformer layer, leading to limited efficiency gains. Notably, when WeiToP is also placed after the second layer with a keep ratio of 0.4, it achieves an average Recall@1 of 69.4 (Fig.~\ref{fig:layer}), outperforming FastV by 4.7\%. Overall, these results demonstrate that WeiToP provides a more flexible and VPR-oriented token pruning strategy, achieving a superior balance between robustness and efficiency across diverse scenarios.

\subsection{Ablation Study}\label{sec:moreablation}
\subsubsection{Different Backbones}
To investigate the scalability of WeiAD and WeiToP, we apply both modules to DINOv2 ViT-S and ViT-L backbones and compare against SALAD under the same settings. The results in Tab.~\ref{tab:diff_size} show that WeiAD consistently outperforms SALAD across backbones of different capacities. Moreover, consistent with the observations on ViT-B, WeiToP achieves near-optimal performance on MSLS-Val and Pitts250k-test with mild token pruning when applied to both ViT-S and ViT-L. Notably, as illustrated in Fig.~\ref{fig:diff_backbone}, the performance gains from WeiToP become more pronounced as the backbone size increases. These results further highlight that larger models offer a higher performance upper bound for VPR, while integrating WeiToP enables a more favorable accuracy–efficiency trade-off during online inference, particularly for high-capacity backbones.

\begin{table}[t]
\caption{Complete results of \textbf{WeiAD} + \textbf{WeiToP}, and our reproduced \textbf{SALAD} in different sizes of DINOv2. The best results without pruning are highlighted in \textbf{bold}. \redbf{Bold red} indicates that the performance after pruning still achieves or surpasses the best results. *The MSLS challenge suspended submissions due to the closure of platform on Jan 20th in 2026, so we can only report partial results.}\label{tab:diff_size}
\resizebox{\columnwidth}{!}{
\begin{tabular}{ccccccccccccc>{\columncolor{gray!10}}c>{\columncolor{gray!10}}c}
\toprule[1.5pt]
\multicolumn{1}{l}{\multirow{2}{*}{\textbf{Method}}}               &  \multirow{2}{*}{$\boldsymbol{\rho}$}   & \textbf{Latency}  & \multicolumn{2}{c}{\textbf{MSLS-val}}  & \multicolumn{2}{c}{\textbf{NordLand}} & \multicolumn{2}{c}{\textbf{Pitts250k-test}} & \multicolumn{2}{c}{\textbf{SPED}} & \multicolumn{2}{c}{\textbf{AmsterTime}} & \multicolumn{2}{>{\columncolor{gray!10}}c}{\textbf{Mean}}   \\ \cline{4-15} 
&                       &  (ms)   & R@1  & R@5   & R@1   & R@5  & R@1    & R@5    & R@1   & R@5   & R@1  & R@5 & R@1  & R@5 \\ \midrule
\rowcolor{gray!15}\multicolumn{15}{l}{DINOv2 ViT-S \quad (\textit{Dim. Size 384 \; Params. 21M})} \\
 SALAD                       & /     & \ms{1.22}{0.03} & \ms{90.1}{0.43} & \ms{95.4}{0.13} &
 \ms{64.9}{3.07} & \ms{81.4}{2.88} &
 \ms{\textbf{94.6}}{0.25} & \ms{98.4}{0.05} &
 \ms{\textbf{88.3}}{0.58} & \ms{93.7}{0.19} &
 \ms{53.5}{0.98} & \ms{74.5}{0.66} &
\ms{78.3}{1.06} & \ms{88.7}{0.78} \\ \hdashline
 WeiAD                       & 1.00  & \ms{1.04}{0.02} &\ms{\textbf{90.7}}{0.31} & \ms{\textbf{95.8}}{0.23} &
 \ms{\textbf{65.4}}{2.33} & \ms{\textbf{81.5}}{1.82} &
 \ms{\textbf{94.6}}{0.07} & \ms{\textbf{98.5}}{0.15} &
 \ms{88.0}{0.84} & \ms{\textbf{93.9}}{0.36} &
 \ms{\textbf{54.0}}{1.00} & \ms{\textbf{75.0}}{0.56} &
\ms{\textbf{78.6}}{0.91} & \ms{\textbf{88.9}}{0.62} \\ \hdashline
 \multirow{9}{*}{\rot{+ WeiToP}}  & 0.95  & \ms{0.97}{0.02} & \ms{\redbf{90.8}}{0.44} & \ms{\redbf{95.9}}{0.23} &
 \ms{64.3}{2.20} & \ms{81.1}{1.55} &
 \ms{94.5}{0.14} & \ms{98.4}{0.14} &
 \ms{87.8}{0.61} & \ms{93.7}{0.21} &
 \ms{53.5}{1.04} & \ms{\redbf{75.0}}{0.70} &
\ms{78.2}{0.89} & \ms{88.8}{0.57} \\ 
                            & 0.90  & \ms{0.93}{0.02} & \ms{\redbf{90.9}}{0.33} & \ms{\redbf{95.9}}{0.29} &
 \ms{64.0}{2.26} & \ms{81.1}{1.49} &
 \ms{94.4}{0.18} & \ms{98.3}{0.13} &
 \ms{87.8}{0.63} & \ms{93.5}{0.28} &
 \ms{53.1}{1.02} & \ms{74.8}{0.83} &
\ms{78.0}{0.88} & \ms{88.7}{0.61} \\
                            & 0.85  & \ms{0.90}{0.02} & \ms{\redbf{90.7}}{0.46} & \ms{\redbf{95.9}}{0.22} &
 \ms{63.8}{2.27} & \ms{80.8}{1.60} &
 \ms{94.2}{0.17} & \ms{98.2}{0.12} &
 \ms{87.4}{0.95} & \ms{93.2}{0.56} &
 \ms{52.5}{1.03} & \ms{74.3}{0.40} &
\ms{77.7}{0.97} & \ms{88.5}{0.58} \\
                             & 0.80  & \ms{0.86}{0.02} & \ms{90.4}{0.44} & \ms{\redbf{96.0}}{0.10} &
 \ms{63.4}{2.12} & \ms{80.6}{1.47} &
 \ms{94.1}{0.06} & \ms{98.2}{0.15} &
 \ms{86.3}{1.03} & \ms{92.9}{0.27} &
 \ms{52.4}{0.95} & \ms{74.2}{0.42} &
\ms{77.3}{0.92} & \ms{88.4}{0.48} \\
                            & 0.75  & \ms{0.83}{0.02} & \ms{90.1}{0.64} & \ms{\redbf{95.8}}{0.18} &
 \ms{63.1}{2.43} & \ms{80.3}{1.51} &
 \ms{93.8}{0.08} & \ms{98.1}{0.17} &
 \ms{85.5}{0.81} & \ms{92.3}{0.22} &
 \ms{51.9}{0.78} & \ms{74.1}{0.40} &
\ms{76.9}{0.95} & \ms{88.1}{0.50} \\
                            & 0.70  & \ms{0.76}{0.03} & \ms{90.0}{0.65} & \ms{95.5}{0.22} &
 \ms{62.4}{2.09} & \ms{79.7}{1.64} &
 \ms{93.4}{0.12} & \ms{98.0}{0.19} &
 \ms{84.1}{1.33} & \ms{91.9}{0.42} &
 \ms{51.1}{1.09} & \ms{73.4}{0.72} &
\ms{76.2}{1.06} & \ms{87.7}{0.64} \\
                            & 0.60  & \ms{0.69}{0.02} & \ms{89.3}{0.36} & \ms{95.4}{0.52} &
 \ms{60.0}{2.51} & \ms{77.8}{1.97} &
 \ms{92.2}{0.15} & \ms{97.5}{0.14} &
 \ms{82.2}{1.69} & \ms{91.1}{0.29} &
 \ms{48.9}{1.09} & \ms{70.8}{0.32} &
\ms{74.5}{1.16} & \ms{86.5}{0.65} \\
                            & 0.50  & \ms{0.63}{0.02} & \ms{88.1}{0.43} & \ms{94.9}{0.19} &
 \ms{54.6}{2.18} & \ms{73.3}{1.89} &
 \ms{90.6}{0.38} & \ms{96.8}{0.06} &
 \ms{79.5}{1.53} & \ms{89.4}{0.58} &
 \ms{44.9}{1.22} & \ms{67.0}{1.16} &
\ms{71.5}{1.15} & \ms{84.3}{0.77} \\
                             & 0.40  & \ms{0.53}{0.02} & \ms{85.3}{0.63} & \ms{93.9}{0.42} &
 \ms{46.1}{2.09} & \ms{64.6}{2.52} &
 \ms{87.7}{0.20} & \ms{95.7}{0.23} &
 \ms{75.7}{0.98} & \ms{86.3}{0.74} &
 \ms{38.2}{0.94} & \ms{60.0}{0.77} &
\ms{66.6}{0.97} & \ms{80.1}{0.93} \\ \midrule
\rowcolor{gray!15}\multicolumn{15}{l}{DINOv2 ViT-B \quad (\textit{Dim. Size 768 \; Params. 86M})} \\
 SALAD                       & /     & \ms{2.41}{0.00} & \ms{91.4}{0.69} & \ms{\textbf{96.2}}{0.28} &
 \ms{71.1}{2.45} & \ms{85.6}{2.03} &
 \ms{94.9}{0.20} & \ms{98.5}{0.20} &
 \ms{90.3}{0.50} & \ms{\textbf{95.4}}{0.36} &
 \ms{58.0}{1.10} & \ms{78.6}{0.98} &
\ms{81.1}{0.99} & \ms{90.9}{0.77} \\ \hdashline
 WeiAD                       & 1.00  & \ms{2.41}{0.00} & \ms{\textbf{91.7}}{0.47} & \ms{\textbf{96.2}}{0.26} &
 \ms{\textbf{73.1}}{2.43} & \ms{\textbf{86.7}}{2.02} &
 \ms{\textbf{95.1}}{0.20} & \ms{\textbf{98.8}}{0.07} &
 \ms{\textbf{90.6}}{0.30} & \ms{\textbf{95.4}}{0.44} &
 \ms{\textbf{58.4}}{0.72} & \ms{\textbf{79.5}}{0.75} &
 \ms{\textbf{81.8}}{0.82} & \ms{\textbf{91.3}}{0.71} \\\hdashline
 \multirow{9}{*}{\rot{+ WeiToP}}  & 0.95  & \ms{2.23}{0.00} & \ms{\redbf{91.8}}{0.60} & \ms{\redbf{96.3}}{0.10} &
 \ms{72.8}{2.22} & \ms{\redbf{86.7}}{1.81} &
 \ms{95.0}{0.23} & \ms{98.6}{0.05} &
 \ms{90.4}{0.18} & \ms{95.3}{0.42} &
 \ms{\redbf{58.4}}{0.88} & \ms{79.4}{0.76} &
\ms{81.7}{0.82} & \ms{91.2}{0.63} \\
                             & 0.90  & \ms{2.13}{0.00} & \ms{\redbf{91.8}}{0.68} & \ms{\redbf{96.3}}{0.17} &
 \ms{71.5}{2.21} & \ms{85.9}{1.97} &
 \ms{94.9}{0.25} & \ms{98.6}{0.03} &
 \ms{90.0}{0.16} & \ms{95.2}{0.34} &
 \ms{57.7}{1.00} & \ms{79.1}{0.70} &
\ms{81.2}{0.86} & \ms{91.0}{0.64} \\
                            & 0.85  & \ms{2.03}{0.01} & \ms{\redbf{91.9}}{0.64} & \ms{\redbf{96.3}}{0.17} &
 \ms{70.8}{2.20} & \ms{85.5}{2.06} &
 \ms{94.8}{0.22} & \ms{98.6}{0.04} &
 \ms{89.6}{0.28} & \ms{95.0}{0.28} &
 \ms{56.9}{0.93} & \ms{78.6}{0.87} &
\ms{80.8}{0.85} & \ms{90.8}{0.68} \\
                            & 0.80  & \ms{1.94}{0.01} & \ms{\redbf{91.8}}{0.46} & \ms{\redbf{96.3}}{0.22} &
 \ms{70.5}{2.03} & \ms{85.1}{2.08} &
 \ms{94.6}{0.25} & \ms{98.5}{0.06} &
 \ms{89.6}{0.54} & \ms{94.7}{0.46} &
 \ms{56.6}{1.24} & \ms{78.2}{0.67} &
\ms{80.6}{0.90} & \ms{90.6}{0.70} \\
                            & 0.75  & \ms{1.83}{0.00} & \ms{91.6}{0.42} & \ms{96.1}{0.14} &
 \ms{70.1}{1.69} & \ms{84.8}{2.08} &
 \ms{94.3}{0.22} & \ms{98.4}{0.05} &
 \ms{89.5}{0.64} & \ms{94.6}{0.45} &
 \ms{56.0}{1.28} & \ms{77.8}{1.04} &
\ms{80.3}{0.85} & \ms{90.3}{0.75} \\
                            & 0.70  & \ms{1.68}{0.01} & \ms{91.3}{0.31} & \ms{96.1}{0.34} &
 \ms{69.5}{1.42} & \ms{84.7}{2.03} &
 \ms{94.0}{0.16} & \ms{98.3}{0.06} &
 \ms{89.2}{0.72} & \ms{94.4}{0.36} &
 \ms{55.3}{1.06} & \ms{76.9}{0.77} &
\ms{79.9}{0.73} & \ms{90.1}{0.71} \\
                           & 0.60  & \ms{1.50}{0.01} & \ms{91.1}{0.61} & \ms{96.0}{0.27} &
 \ms{67.6}{1.49} & \ms{83.6}{1.64} &
 \ms{93.1}{0.23} & \ms{98.0}{0.09} &
 \ms{87.5}{0.43} & \ms{93.8}{0.30} &
 \ms{53.0}{1.08} & \ms{75.0}{0.63} &
\ms{78.5}{0.77} & \ms{89.3}{0.58} \\
                            & 0.50  & \ms{1.30}{0.00} & \ms{90.4}{0.73} & \ms{95.9}{0.30} &
 \ms{62.3}{1.43} & \ms{80.0}{1.52} &
 \ms{91.5}{0.39} & \ms{97.4}{0.12} &
 \ms{85.5}{0.44} & \ms{92.5}{0.40} &
 \ms{48.7}{1.20} & \ms{70.6}{1.67} &
\ms{75.7}{0.84} & \ms{87.3}{0.80} \\
                            & 0.40  & \ms{1.09}{0.00} & \ms{87.7}{0.88} & \ms{94.8}{0.47} &
 \ms{53.3}{1.70} & \ms{71.0}{1.55} &
 \ms{88.2}{0.50} & \ms{96.2}{0.22} &
 \ms{82.4}{1.22} & \ms{90.9}{0.60} &
 \ms{41.8}{1.71} & \ms{64.0}{1.68} &
\ms{70.7}{1.20} & \ms{83.4}{0.90} \\ \midrule
\rowcolor{gray!15}\multicolumn{15}{l}{DINOv2 ViT-L \quad (\textit{Dim. Size 1024 \; Params. 300M})} \\
 SALAD                       & /     & \ms{7.01}{0.06} & \ms{\textbf{92.7}}{0.58} & \ms{\textbf{96.8}}{0.15} &
 \ms{74.6}{2.60} & \ms{88.2}{1.82} &
 \ms{\textbf{95.8}}{0.11} & \ms{\textbf{98.9}}{0.06} &
 \ms{92.8}{0.15} & \ms{96.9}{0.21} &
 \ms{62.9}{0.81} & \ms{82.9}{0.55} &
\ms{83.8}{0.85} & \ms{92.7}{0.56} \\ \hdashline
 WeiAD                       & 1.00  & \ms{7.10}{0.01} & \ms{\textbf{92.7}}{0.61} & \ms{96.7}{0.49} &
 \ms{\textbf{75.6}}{1.46} & \ms{\textbf{89.0}}{1.39} &
 \ms{\textbf{95.8}}{0.28} & \ms{\textbf{98.9}}{0.15} &
 \ms{\textbf{93.1}}{0.39} & \ms{\textbf{97.0}}{0.24} &
 \ms{\textbf{63.4}}{0.42} & \ms{\textbf{83.7}}{0.19} &
\ms{\textbf{84.1}}{0.63} & \ms{\textbf{93.1}}{0.49} \\ \hdashline
 \multirow{9}{*}{\rot{+ WeiToP}}  & 0.95  & \ms{6.62}{0.02} & \ms{\redbf{92.7}}{0.66} & \ms{\redbf{96.7}}{0.33} &
 \ms{73.7}{1.57} & \ms{86.8}{1.31} &
 \ms{\redbf{95.9}}{0.28} & \ms{\redbf{99.0}}{0.15} &
 \ms{92.5}{0.13} & \ms{96.7}{0.39} &
 \ms{62.3}{0.49} & \ms{83.6}{0.23} &
\ms{83.4}{0.63} & \ms{92.6}{0.48} \\
                             & 0.90  & \ms{6.32}{0.02} & \ms{92.5}{0.67} & \ms{\redbf{96.8}}{0.31} &
 \ms{73.3}{1.43} & \ms{86.5}{1.28} &
 \ms{\redbf{95.8}}{0.23} & \ms{\redbf{98.9}}{0.13} &
 \ms{92.4}{0.30} & \ms{96.5}{0.19} &
 \ms{62.6}{0.47} & \ms{83.4}{0.35} &
\ms{83.3}{0.62} & \ms{92.4}{0.45} \\
                            & 0.85  & \ms{6.01}{0.05} & \ms{92.6}{0.58} & \ms{\redbf{96.8}}{0.26} &
 \ms{72.6}{1.21} & \ms{86.0}{1.22} &
 \ms{95.6}{0.23} & \ms{\redbf{98.9}}{0.13} &
 \ms{92.2}{0.22} & \ms{96.3}{0.34} &
 \ms{62.3}{0.43} & \ms{83.2}{0.44} &
\ms{83.1}{0.53} & \ms{92.3}{0.48} \\
                             & 0.80  & \ms{5.68}{0.01} & \ms{92.5}{0.49} & \ms{\redbf{96.8}}{0.23} &
 \ms{71.9}{1.12} & \ms{85.7}{1.14} &
 \ms{95.4}{0.24} & \ms{98.8}{0.14} &
 \ms{92.0}{0.63} & \ms{96.2}{0.43} &
 \ms{61.7}{0.54} & \ms{83.5}{0.22} &
\ms{82.7}{0.60} & \ms{92.2}{0.43} \\
                            & 0.75  & \ms{5.38}{0.02} & \ms{92.3}{0.69} & \ms{\redbf{96.8}}{0.20} &
 \ms{71.1}{1.12} & \ms{85.0}{1.05} &
 \ms{95.2}{0.26} & \ms{98.8}{0.13} &
 \ms{91.3}{0.08} & \ms{95.8}{0.47} &
 \ms{61.4}{0.63} & \ms{82.7}{0.38} &
\ms{82.3}{0.56} & \ms{91.8}{0.45} \\
                            & 0.70  & \ms{4.92}{0.02} & \ms{92.3}{0.48} & \ms{\redbf{96.7}}{0.24} &
 \ms{69.9}{1.09} & \ms{84.0}{0.97} &
 \ms{95.0}{0.28} & \ms{98.7}{0.18} &
 \ms{90.7}{0.56} & \ms{95.4}{0.62} &
 \ms{60.9}{0.68} & \ms{82.3}{0.43} &
\ms{81.8}{0.62} & \ms{91.4}{0.49} \\
                           & 0.60  & \ms{4.32}{0.02} & \ms{91.8}{0.49} & \ms{96.4}{0.30} &
 \ms{66.8}{1.25} & \ms{81.5}{1.05} &
 \ms{94.0}{0.44} & \ms{98.5}{0.18} &
 \ms{89.4}{0.68} & \ms{95.1}{0.36} &
 \ms{58.6}{0.79} & \ms{79.9}{1.40} &
\ms{80.1}{0.73} & \ms{90.3}{0.66} \\
                            & 0.50  & \ms{3.69}{0.02} & \ms{91.1}{0.46} & \ms{96.4}{0.18} &
 \ms{60.2}{1.82} & \ms{75.7}{1.63} &
 \ms{92.5}{0.65} & \ms{98.0}{0.23} &
 \ms{87.7}{1.04} & \ms{94.3}{0.48} &
 \ms{54.3}{0.73} & \ms{76.4}{0.93} &
\ms{77.2}{0.94} & \ms{88.2}{0.69} \\
                             & 0.40  &\ms{2.97}{0.01} & \ms{89.4}{0.46} & \ms{95.8}{0.16} &
 \ms{50.4}{2.03} & \ms{66.2}{1.89} &
 \ms{89.7}{0.59} & \ms{97.1}{0.31} &
 \ms{84.5}{1.09} & \ms{93.0}{0.37} &
 \ms{46.5}{0.54} & \ms{69.4}{1.42} &
\ms{72.1}{0.94} & \ms{84.3}{0.83} \\ \bottomrule[1.5pt]
\end{tabular}
}
\end{table}

\subsubsection{Different Layers}\label{sec:difflayer}
\begin{figure}[h]
\centering
\includegraphics[width=0.5\linewidth]{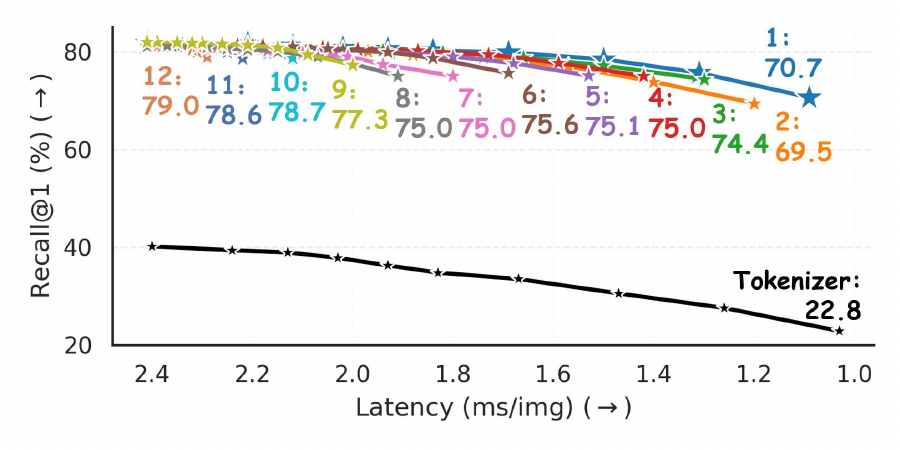}
\caption{Efficiency-accuracy trade-off of the \textbf{WeiToP} pruning module located after different layers of ViT and the initial tokenizer.}
\label{fig:tokenizer}
\end{figure}

We further investigate the effect of placing the pruning module at different transformer layers. For each placement, we conduct five repeated runs and report averaged results across datasets, excluding MSLS-C for consistency. As shown in Fig.~\ref{fig:layer}, placing the pruning module immediately after the first transformer layer achieves the most favorable accuracy–efficiency trade-off in ViT backbones. We additionally explore inserting the pruning module before the transformer, \ie, directly after the initial patch embedding. In this case, VPR performance degrades sharply, and joint learning further interferes with the full-retention performance of WeiAD, as illustrated in Fig.~\ref{fig:tokenizer}. We report the Recall@1 when $\rho$ is 0.4. These results indicate that excessively early pruning disrupts the formation of stable and discriminative visual representations.

\subsubsection{Choice of Hyperparameters}\label{sec:choixeofhp}
Following SALAD, we adopt a set of common hyperparameters for all experiments. We fine-tune only the last 4 transformer layers. Each training batch contains 60 locations from the GSV-Cities dataset~\cite{GSV-Cities}, with 4 images per location. We use the AdamW optimizer~\cite{Adam} with an initial learning rate of $6\times10^{-5}$ and a linear decay schedule. We set most hyperparameters following prior work and empirical validation.

Some key hyperparameters are further analyzed. We first study the distribution of cluster tiers and find that a head-heavy configuration $\{24, 20, 16, 4\}$ performs best, while tail-heavy or uniform distributions lead to inferior results, as reported in Tab.~\ref{tab:distributiontiers}. 

\begin{table}[h]
\caption{Distribution of clusters $\{c_0, c_1, \dots, c_{T^{(w)}-1}\}$ in each tier.}\label{tab:distributiontiers}
\centering
\resizebox{\linewidth}{!}{
\begin{tabular}{c:c:c:c:>{\columncolor{gray!10}}c>{\columncolor{gray!10}}c>{\columncolor{gray!10}}c>{\columncolor{gray!10}}c}
\toprule[1.5pt]
\multicolumn{4}{c}{\textbf{Cluster Number} of Each Tier (Important $\rightarrow$ Unimportant)}    &   \multicolumn{4}{>{\columncolor{gray!10}}c}{\textbf{Mean} \text{\scriptsize w/o MSLS-C}}            \\ \hline
 $\exp(\theta_0) $& $\exp(\theta_0) -  \operatorname{softplus}(\theta_\Delta)$&  $\exp(\theta_0) - 2 \cdot  \operatorname{softplus}(\theta_\Delta)$& $\exp(\theta_0) - 3 \cdot \operatorname{softplus}(\theta_\Delta))$& R@1      & R@5   & R@10      & R@20   \\ \midrule
4&16&20&24& \ms{81.0}{0.81} & \ms{91.2}{0.73}  & \ms{93.1}{0.67} & \ms{95.0}{0.63} \\ 
16&16&16&16& \ms{81.2}{0.74} & \ms{91.0}{0.69} & \ms{\textbf{93.5}}{0.62} & \ms{\textbf{95.2}}{0.68} \\
24&20&16&4& \ms{\textbf{81.8}}{0.82} & \ms{\textbf{91.3}}{0.71} & \ms{93.4}{0.71} & \ms{\textbf{95.2}}{0.48} \\\bottomrule[1.5pt]
\end{tabular}
}
\end{table}

\begin{figure}[t]
\centering
\includegraphics[width=0.5\linewidth]{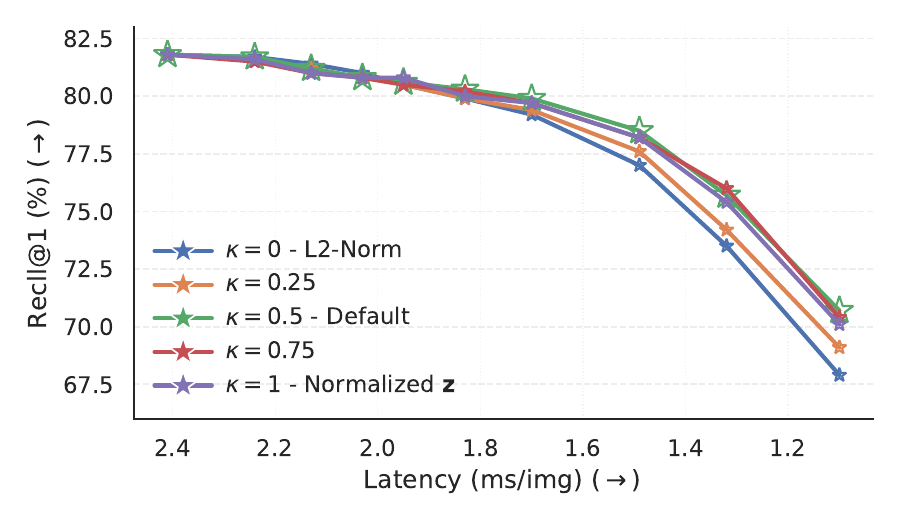}
\caption{Efficiency-accuracy trade-off with different balancing coefficient $\kappa$ for token-level importance scores.}
\label{fig:diffkappa}
\end{figure}

We also analyze the balancing coefficient $\kappa$ between the learned importance scores and token feature magnitudes. Empirically, $\kappa=0.5$ achieves the best performance, as shown in Fig.~\ref{fig:diffkappa}. Using only the normalized predicted importance logits from WeiToP (\ie, $\kappa=1$) still outperforms relying solely on the $\ell_2$ norm of token features ($\kappa=0$) and remains close to the best performance. A relatively high weight, such as $\kappa=0.75$, also yields competitive results, whereas a lower value ($\kappa=0.25$) leads to a noticeable performance degradation.

\begin{figure}[t]
\centering
\includegraphics[width=0.5\linewidth]{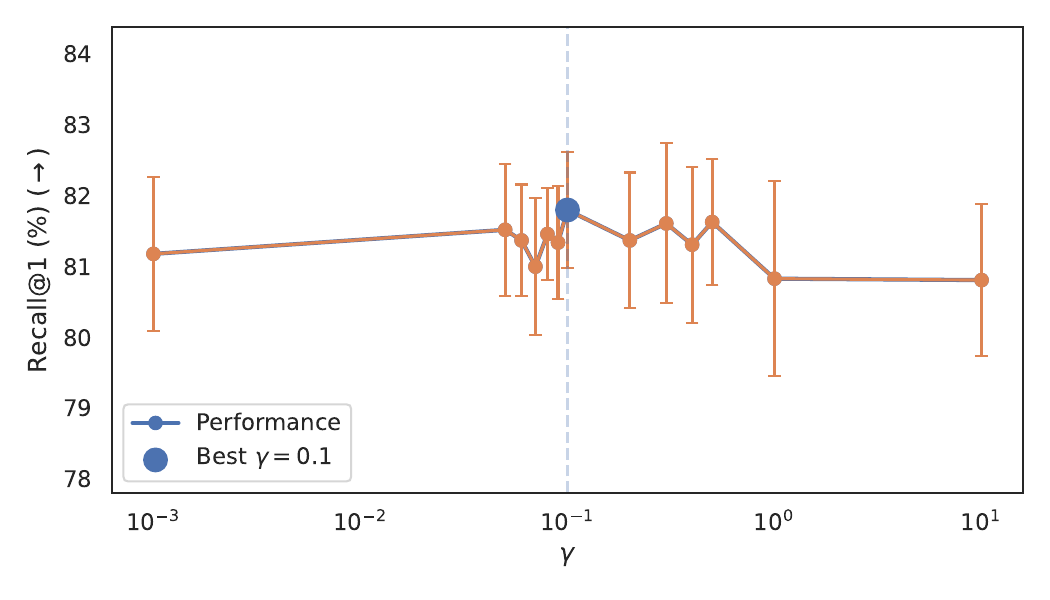}
\caption{Performance of WeiAD under different magnitudes of $\gamma$ during joint learning.}
\label{fig:gamma}
\end{figure}

Finally, we explore the weight $\gamma$ of the distillation loss and find that $\gamma=0.1$ is optimal, as illustrated in Fig.~\ref{fig:gamma}.

\begin{figure}[t]
\centering
\includegraphics[width=\linewidth]{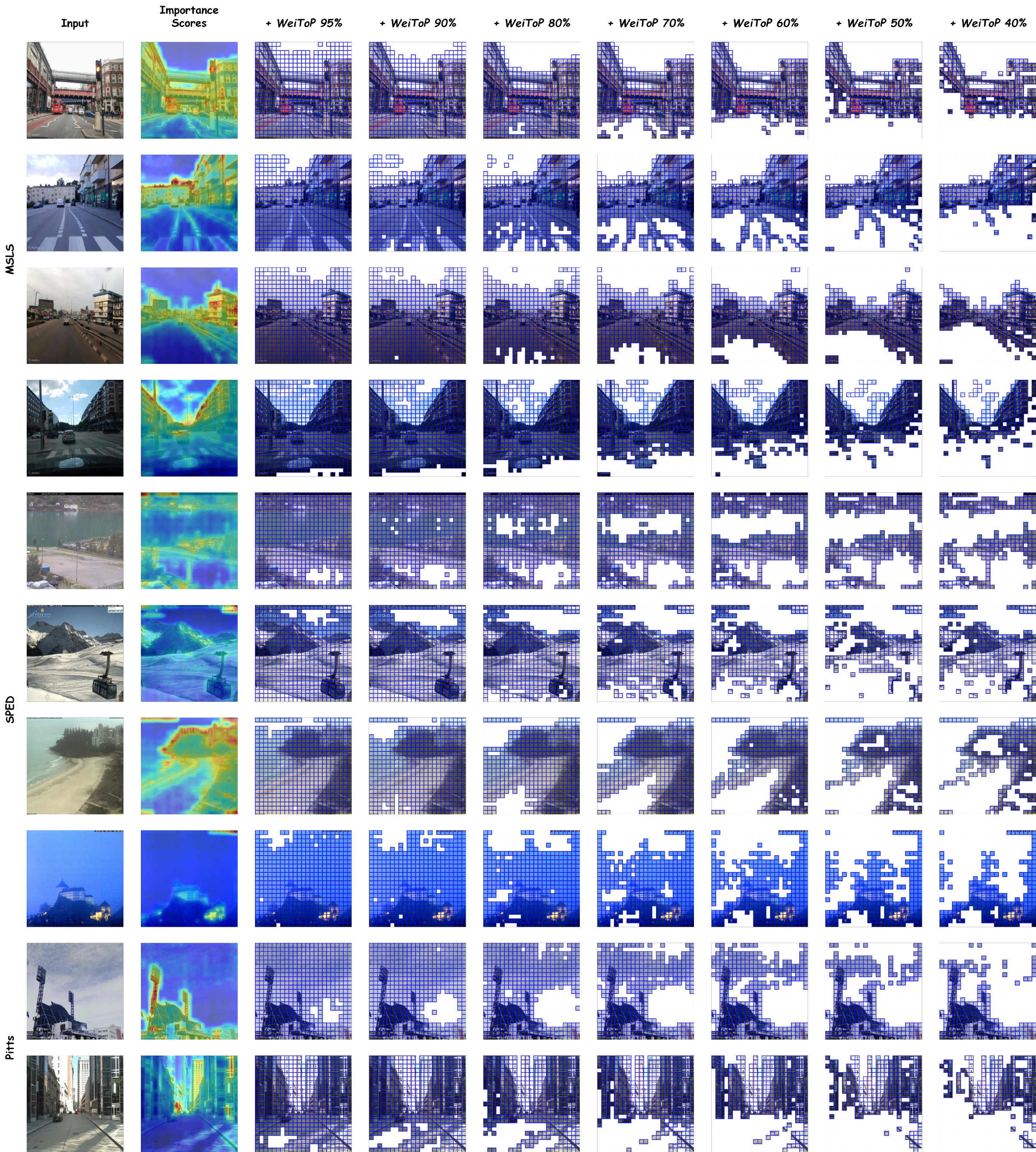}
\end{figure}

\section{Limitations and Future Work}

Despite its effectiveness, our approach has several limitations that suggest promising directions for future research. First, our weighted aggregation relies on a fixed cluster structure learned during training. Although empirical results demonstrate that clusters exhibit stable semantic and spatial biases, the optimal number of clusters and their partition into tiers are still treated as hyperparameters. Automatically adapting the cluster structure or learning the weighting hierarchy in a data-driven or task-adaptive manner could further improve robustness and generalization. Second, token importance in our framework is estimated primarily from early-layer features. While this design enables efficient pruning and flexible inference-time control, early representations may lack high-level semantic context in extremely challenging scenarios, such as severe viewpoint or appearance changes. Incorporating multi-layer aggregated importance signals, while maintaining efficiency, remains an open problem. Finally, although our method generalizes well across different Vision Transformer backbones, its current formulation is primarily designed for transformer-based architectures. Exploring analogous weighting and pruning mechanisms for other representation paradigms, such as hybrid pruning-merging models~\cite{ToFu} or emerging vision foundation models like DINOv3~\cite{DINOV3}, is a promising direction for future work. Beyond these directions, we plan to further extend the proposed self-distilled pruning framework to a broader range of architectures and investigate hybrid strategies that combine token pruning with feature- or descriptor-level fusion. Another important avenue is to integrate our framework into multi-stage retrieval pipelines~\cite{R2Former,SelaVPR}, segment-level VPR systems~\cite{SegVLAD,MuSSel-V}, and cross-view geolocalization~\cite{zhu2022transgeo,zheng2026rho}, as well as to explore its impact on large-scale visual localization, SLAM, and other downstream applications.

%


\end{document}